%% file: AnonymousSubmission2027.tex
\documentclass[letterpaper]{article} 
\usepackage{aaai2027}  
\nocopyright
\usepackage[hyphens]{url}  
\usepackage{graphicx} 
\usepackage{natbib}  
\usepackage{caption} 
\usepackage{amsmath}
\usepackage{multirow}
\usepackage{tabularx}
\usepackage{array}
\usepackage{algorithm}

\usepackage{amssymb}

\usepackage{algpseudocode}

\usepackage{booktabs}

\usepackage{algpseudocode}
\usepackage{newfloat}
\usepackage{listings}
\DeclareCaptionStyle{ruled}{labelfont=normalfont,labelsep=colon,strut=off} 
\floatstyle{ruled}
\newfloat{listing}{tb}{lst}{}
\floatname{listing}{Listing}

\usepackage{booktabs}

\title{MOAE: Multi-Objective Agent Evolution with Pareto-Preserving Search}

\author{
Hengle Jiang, Qijun Cai, Ziying Luo, Ke Tang\corresponding
}
\affiliations{
	Guangdong Provincial Key Laboratory of Brain-inspired Intelligent Computation\\
    Department of Computer Science and Engineering, Southern University of Science and Technology\\

{\footnotesize \texttt{\{jianghl2025, caiqj2026, luozy2025\}@mail.sustech.edu.cn} \qquad \texttt{tangk3@sustech.edu.cn}}
}

\begin{document}

\maketitle

\begin{abstract}
As LLM-based agents continue to advance, their evaluation has become increasingly multifaceted: a capable agent must not only achieve high task completion accuracy but also perform well in interaction quality, safety, and efficiency, raising a central question: can these objectives be optimized simultaneously? Existing methods have considered multiple objectives, but many collapse heterogeneous measurements into a fixed scalar score. Such scalarization depends on metric normalization and preference weights and may discard candidates that represent useful deployment trade-offs. We introduce Multi-Objective Agent Evolution (MOAE), which organizes iterative in-context refinement as a Pareto-preserving evolutionary search over complete agent rollouts. Given a limited rollout budget, MOAE maintains an empirical archive of non-dominated candidates, uses objective-specific diagnostics to guide offspring generation, and applies constraint-aware selection only at deployment. This separates candidate preservation during search from the preference used to return a final solution. The procedure requires no parameter updates and allows each objective to be replaced by any measurable property, which we instantiate as task performance, trajectory quality, and safety. Experiments on TravelPlanner and AgentDojo show that MOAE consistently improves task performance and trajectory quality while maintaining strong safety under matched rollout budgets. Search-behavior analysis further shows that Pareto preservation expands the attainable objective region and increases the frequency of joint improvement. These results demonstrate the potential of Pareto-preserving in-context evolution for optimizing multiple agent properties without committing to a fixed scalarization during search.
\end{abstract}

\input{sections/introduction}
\input{sections/related}
\input{sections/method}
\input{sections/experiments}

\input{sections/conclusion}

\bibliography{aaai2027}

\clearpage
\input{sections/appendix}

\end{document}

%% file: sections/introduction.tex
\section{Introduction}
\label{sec:intro}

Large language model based agents have rapidly evolved from text generators into interactive systems that plan, invoke tools, process environmental observations, and revise their decisions over multiple steps \cite{yao2022react,xie2024travelplanner}. As their capabilities expand, evaluating them solely by final answer accuracy is no longer sufficient. A useful agent should complete the requested task while producing a valid and well-grounded interaction trajectory, resisting unsafe instructions, and controlling its operational cost. Recent benchmarks accordingly evaluate planning constraints, tool-use behavior, trajectory quality, efficiency, and robustness \cite{liu2024agentbench,ma2024agentboard,jiang-tang-2026-agents, zhan2024injecagent,deb2024agentdojo}. These dimensions are related but not always aligned. Additional tool use may improve evidence coverage while introducing redundant or invalid actions, whereas conservative behavior may reduce safety risk but prevent useful exploration. This motivates a central question: can several properties of an LLM agent be improved jointly rather than optimized in isolation?

Existing methods address different parts of this question. Iterative refinement and verbal feedback revise individual responses or trajectories \cite{madaan2023selfrefine,shinn2023reflexion}, while prompt and program optimizers search for reusable instructions over a development set \cite{yang2024opro,guo2024evoprompt,fernando2024promptbreeder,opsahlong2024mipro,agrawal2025gepa}. Alignment and agent-optimization methods have also considered multiple preferences, capability and safety, or performance and cost \cite{zhou2024modpo,zhong2024panacea,yuan2025evoagent,rosser2025agentbreeder}. These studies establish that agent optimization need not be limited to one objective. However, their trade-offs are commonly encoded through fixed scalarization, predefined constraints, learned preference parameters, or persistent artifacts optimized across a development set. Less attention has been given to limited-budget, per-query optimization that preserves several rollout-level trade-offs during search and applies the final preference only when selecting the deployed response.

This setting presents three practical challenges. First, agent objectives are often non-differentiable, measured on different scales, and observable only after a complete interaction. Second, improvements may conflict, so retaining a single incumbent can discard useful trade-offs. Fixed scalarization also requires predefined normalization and preference weights; poor choices can select an undesirable compromise or eliminate candidates valuable under another deployment requirement. Third, independent sampling provides diversity but does not use observed failures to guide later rollouts. An effective online optimizer should therefore preserve candidates with different strengths, target specific weaknesses, and defer the final preference to deployment.

We propose Multi-Objective Agent Evolution (MOAE), a Pareto-preserving framework for limited-budget, per-query agent optimization. MOAE keeps the model, tools, and agent architecture fixed while evolving a compact execution policy. Starting from a base rollout, it evaluates complete trajectories, maintains an empirical archive of locally non-dominated candidates, selects objective-specific parents, and uses evaluator diagnostics to guide in-context mutation. Candidate preservation is separated from constraint-aware deployment, allowing the final preference to be applied after search. We instantiate MOAE with task performance, trajectory quality, and safety risk. Experiments on TravelPlanner and AgentDojo show improvements over the unoptimized agent and competitive balanced solutions under matched rollout budgets.

This work makes three contributions. First, we formulate limited-budget, per-query agent optimization as a multi-objective problem that separates rollout objectives, candidate preservation, deployment preferences, and search cost. Second, we introduce an evolutionary procedure combining empirical Pareto preservation, objective-specific parent selection, diagnostic-guided mutation, and constraint-aware deployment without model or architecture updates. Third, we conduct matched-budget evaluations on TravelPlanner and AgentDojo across three language models and multiple search and agent-optimization baselines. The experiments examine both final performance and search behavior, while ablations isolate the effects of mutation, diagnostics, and the non-dominated archive.

%% file: sections/related.tex
\section{Related Work}
\label{sec:related}

\subsection{LLM Agent Evaluation}
LLM agents combine language models with planning, memory, and external tools, making final-answer accuracy insufficient for characterizing their behavior. AgentBench and AgentBoard evaluate task completion and interaction progress across diverse environments \cite{liu2024agentbench,ma2024agentboard}, while TravelPlanner measures long-horizon planning under transportation, temporal, budget, and commonsense constraints \cite{xie2024travelplanner}. ToolLLM evaluates API selection and function calling, whereas InjecAgent and AgentDojo examine whether agents maintain task utility when tool observations contain indirect prompt injections \cite{qin2023toolllm,zhan2024injecagent,deb2024agentdojo}. These benchmarks establish task performance, trajectory validity, operational efficiency, and safety as distinct aspects of agent quality. However, they primarily provide evaluation protocols and do not specify how several measured properties should be improved jointly.

\subsection{Pairwise Objective Optimization in LLMs and Agents}
Most existing approaches that consider more than task accuracy focus on a predefined pair of objectives. Safe RLHF and Bi-Factorial Preference Optimization balance helpfulness against harmlessness through constrained reinforcement learning or preference optimization \cite{dai2023saferlhf,zhang2025bfpo}, while MODPO and Panacea learn parameterized trade-offs among alignment preferences \cite{zhou2024modpo,zhong2024panacea}. A separate line of work studies accuracy and inference efficiency through reasoning compression, routing, or latency-aware test-time scaling \cite{xia2025tokenskip,yan2025macc,wang2025latencytts}. For tool-integrated agents, ParetoPO optimizes task accuracy together with tool-use efficiency through multi-objective policy training \cite{li2026paretopo}. These methods demonstrate that language models and agents have previously been optimized for multiple objectives. Their operating points, however, are commonly determined through predefined constraints, learned preference parameters, fixed scalarization, or objective-specific optimization structures. MOAE studies a complementary setting in which heterogeneous objectives are observed after complete agent rollouts, empirical non-dominated candidates are preserved under a limited per-query budget, and the final preference is applied at deployment rather than used to collapse the search into one score.



\subsection{LLM Agent Evolution}

Evolutionary methods search over discrete language components without requiring differentiable objectives. EvoPrompt and PromptBreeder evolve task and mutation prompts, while MIPRO searches over instructions and demonstrations for language-model programs \cite{guo2024evoprompt,fernando2024promptbreeder,opsahlong2024mipro}. GEPA uses execution traces and natural-language reflection to optimize reusable textual components, with Pareto-based selection preserving improvements across development examples \cite{agrawal2025gepa}. Evolution has also been extended to agent systems: EvoAgent and AgentBreeder search over reusable multi-agent architectures or scaffolds, with AgentBreeder considering both capability and safety \cite{yuan2025evoagent,rosser2025agentbreeder}. MOCHA optimizes persistent structured skills under correctness and platform constraints, while SkillMOO evolves software-engineering skill bundles according to task performance and inference cost \cite{tanjim2026mocha,gong2026skillmoo}. These methods primarily optimize reusable artifacts across development tasks. In contrast, MOAE performs limited-budget, per-query optimization of a fixed agent, preserves trade-offs among rollout-level objectives, and separates candidate search from deployment selection. A structured comparison is provided in the Appendix.

%% file: sections/method.tex
\begin{figure*}[t]
    \centering
    \includegraphics[width=2.1\columnwidth]{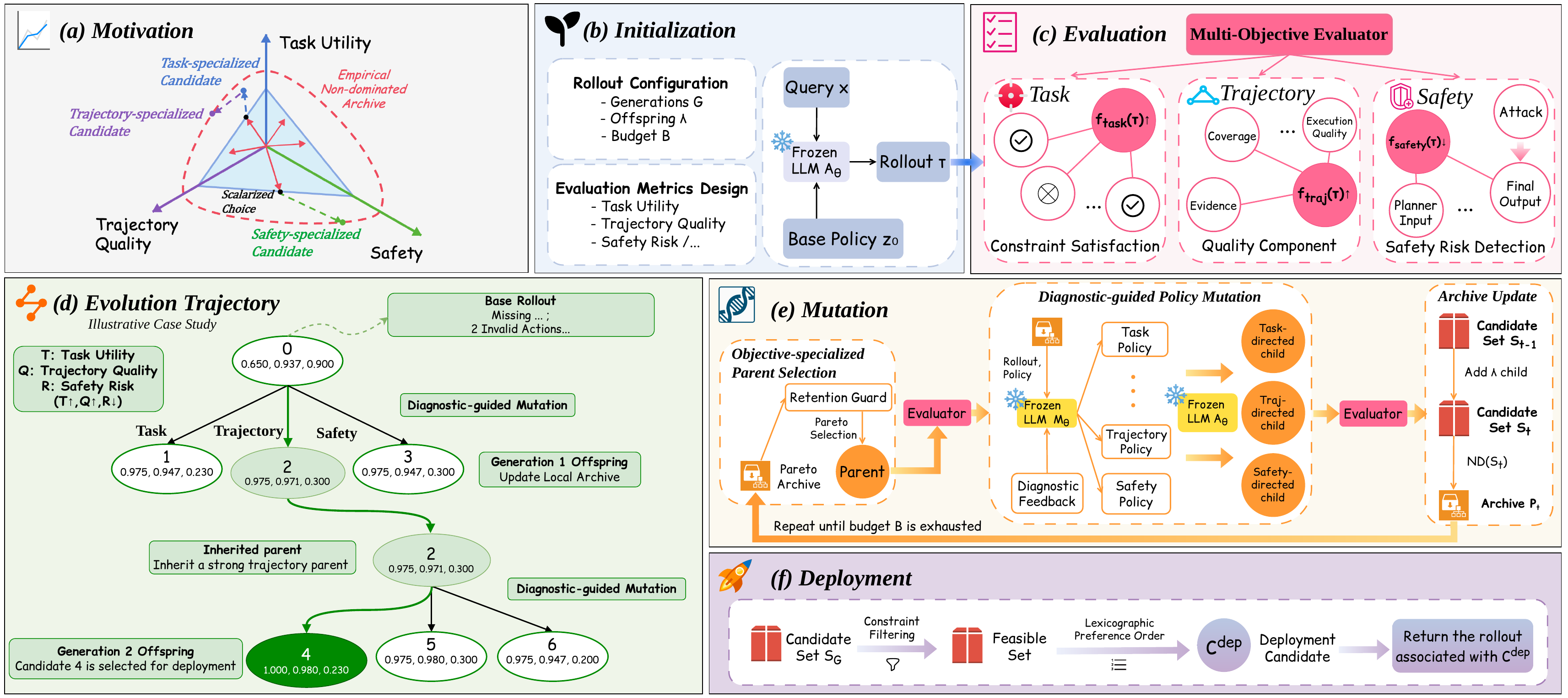}
    \caption{
Overview of MOAE.
(a) Motivation for Pareto preservation.
(b) Rollout initialization.
(c) Multi-objective evaluation.
(d) Illustrative evolution trajectory.
(e) Diagnostic-guided mutation and archive update.
(f) Constraint-aware deployment.
}
\label{fig:moae-overview}
\end{figure*}

\section{Method}
\label{sec:method}

We propose Multi-Objective Agent Evolution (MOAE), a Pareto-preserving framework that organizes per-query refinement as a structured multi-objective search. Figure~\ref{fig:moae-overview} illustrates its complete workflow. MOAE treats the optimization of an LLM agent as a small-scale evolutionary search over its execution policies. Starting from a default policy, the agent performs a limited number of complete rollouts, each of which is evaluated along multiple objectives. Rather than assigning every candidate a fixed aggregate score, MOAE preserves empirical non-dominated candidates with different strengths and uses objective-specific diagnostics to generate compact offspring policies. After the rollout budget is exhausted, a constraint-aware selection rule applies the deployment preference and returns one candidate. The procedure operates through model inference and in-context feedback without updating the parameters of either the agent or the mutation model.
\subsection{Problem Formulation}

Let \(x\in\mathcal{X}\) denote an input query and let
\(\mathcal{A}_{\theta}\) be an LLM-based agent with frozen parameters
\(\theta\). The behavior of the agent is controlled by an execution policy
\(z\in\mathcal{Z}\), which specifies how the agent plans, invokes tools,
verifies intermediate results, and terminates its execution. Given execution
randomness \(\xi\), the agent produces a complete rollout
\begin{equation}
\tau=\mathcal{A}_{\theta}(x;z,\xi).
\end{equation}
A candidate is defined as
\begin{equation}
c=(z,\tau,\mathbf{f}(c),\mathbf{e}(c)),
\end{equation}
where \(\mathbf{f}(c)=[f_1(c),\ldots,f_m(c)]\) and
\(\mathbf{e}(c)\) denote the objective values and evaluator diagnostics
computed from the rollout \(\tau\) stored in \(c\). The diagnostics may include constraint violations, unsupported
decisions, invalid actions, repeated tool calls, or other objective-relevant
signals.

Different objectives may have different optimization directions. We associate
each objective with \(s_j\in\{+1,-1\}\), where \(s_j=+1\) denotes maximization
and \(s_j=-1\) denotes minimization, and define the direction-aligned objective
\begin{equation}
u_j(c)=s_j f_j(c).
\end{equation}
All aligned objectives are therefore maximized. Candidate \(c_a\) Pareto-dominates
candidate \(c_b\), denoted by \(c_a\succ c_b\), if
\begin{equation}
\left[\forall j,\;u_j(c_a)\geq u_j(c_b)\right]
\land
\left[\exists j,\;u_j(c_a)>u_j(c_b)\right].
\end{equation}
For a candidate set \(\mathcal{S}\), its empirical non-dominated subset is
\begin{equation}
\operatorname{ND}(\mathcal{S})
=
\left\{
c\in\mathcal{S}:
\nexists c'\in\mathcal{S}\ \text{s.t.}\ c'\succ c
\right\}.
\end{equation}
Our goal is to improve the agent under a limited online rollout budget while
preserving candidates that represent different objective trade-offs. The
framework only requires each objective to be computable from a completed
rollout and does not assume a particular task, agent architecture, or objective
definition.

\subsection{Multi-Objective Agent Evolution}

MOAE organizes iterative in-context refinement as a multi-objective evolutionary
search. It maintains an empirical Pareto archive, selects objective-specialized
parents, uses evaluator diagnostics to mutate their execution policies, and
evaluates every offspring through a fresh agent rollout. Neither the execution
agent nor the mutation model is updated during this process.

\paragraph{Initialization.}
MOAE first evaluates the default execution policy \(z_0\):
\begin{equation}
\tau_0=\mathcal{A}_{\theta}(x;z_0,\xi_0),
\qquad
c_0=(z_0,\tau_0,\mathbf{f}(c_0),\mathbf{e}(c_0)).
\end{equation}
The evaluated set and Pareto archive are initialized as
\begin{equation}
\mathcal{S}_0=\{c_0\},
\qquad
\mathcal{P}_0=\operatorname{ND}(\mathcal{S}_0).
\end{equation}
With one offspring generated for each objective in every generation, \(G\)
generations require
\begin{equation}
B=1+mG
\end{equation}
complete agent rollouts, including the initial rollout.

\paragraph{Objective-specialized parent selection.}
At generation \(t\), the archive ideal point is defined as
\begin{equation}
\mathbf{u}^{*}_{t}
=
\left[
\max_{c\in\mathcal{P}_{t-1}}u_1(c),
\ldots,
\max_{c\in\mathcal{P}_{t-1}}u_m(c)
\right].
\end{equation}
When an application has a primary utility objective \(q\), MOAE prevents
auxiliary-objective mutations from selecting candidates with substantially
degraded primary utility. It constructs the admissible archive
\begin{equation}
\mathcal{P}^{\mathrm{adm}}_{t-1}
=
\left\{
c\in\mathcal{P}_{t-1}:
u_q(c)
\geq
\max_{a\in\mathcal{P}_{t-1}}u_q(a)-\delta
\right\},
\end{equation}
where \(\delta\geq0\) is a retention tolerance. If no primary objective is
specified, denoted by \(q=\varnothing\), we directly set
\(\mathcal{P}^{\mathrm{adm}}_{t-1}=\mathcal{P}_{t-1}\).

For objective \(j\), the parent search set is
\begin{equation}
\mathcal{R}_{t,j}
=
\begin{cases}
\mathcal{P}_{t-1},
& q=\varnothing\ \text{or}\ j=q,\\
\mathcal{P}^{\mathrm{adm}}_{t-1},
& \text{otherwise},
\end{cases}
\end{equation}
and an objective-specialized parent is selected by
\begin{equation}
p_{t,j}
\in
\arg\max_{c\in\mathcal{R}_{t,j}}u_j(c).
\end{equation}
Deterministic secondary objectives are used to resolve ties. This mechanism
allows archive members with different strengths to serve as parents while
limiting degradation of the primary capability when one is specified.


\paragraph{Diagnostic-guided policy mutation.}
For each selected parent, the evaluator constructs objective-specific diagnostic
feedback
\begin{equation}
r_{t,j}
=
\mathcal{D}_{j}
\left(
\tau(p_{t,j}),
\mathbf{u}(p_{t,j}),
\mathbf{u}^{*}_{t},
\mathbf{e}(p_{t,j})
\right).
\end{equation}
The diagnostic identifies behaviors related to objective \(j\) that can be
improved while exposing the parent objective vector and the current archive
ideal point. A frozen mutation model \(\mathcal{M}_{\phi}\) then generates a
new execution policy:
\begin{equation}
\widetilde{z}_{t,j}
=
\mathcal{M}_{\phi}
\left(
x,
z(p_{t,j}),
\tau(p_{t,j}),
\mathbf{u}(p_{t,j}),
\mathbf{u}^{*}_{t},
r_{t,j},
j
\right).
\end{equation}
The mutation output is passed through a deterministic validator
\(\mathcal{V}\):
\begin{equation}
z_{t,j}
=
\mathcal{V}(\widetilde{z}_{t,j}).
\end{equation}
The validator enforces a compact policy schema and bounds execution parameters.
In our implementation, the resulting policy contains concise planning,
tool-use, verification, and stopping rules. Importantly, the detailed evaluator
feedback is used only by the mutation model. The execution agent receives the
validated compact policy rather than the parent trajectory or evaluator
annotations.

The offspring is evaluated through a fresh complete rollout:
\begin{equation}
\tau_{t,j}
=
\mathcal{A}_{\theta}(x;z_{t,j},\xi_{t,j})
\end{equation}
\begin{equation}
c_{t,j}
=
\left(
z_{t,j},
\tau_{t,j},
\mathbf{f}(c_{t,j}),
\mathbf{e}(c_{t,j})
\right).
\end{equation}
This separation is important because the mutation model proposes a behavioral
change, while the objective values are determined only by executing and
evaluating the resulting agent policy.

\paragraph{Archive update.}
Let
\begin{equation}
\mathcal{C}_t=\{c_{t,1},\ldots,c_{t,m}\}
\end{equation}
denote the offspring generated at generation \(t\). MOAE updates the evaluated
set and archive as
\begin{equation}
\mathcal{S}_t=\mathcal{S}_{t-1}\cup\mathcal{C}_t,
\qquad
\mathcal{P}_t=\operatorname{ND}(\mathcal{P}_{t-1}\cup\mathcal{C}_t).
\end{equation}
An optional crowding-distance rule can be applied if the archive exceeds a
predefined capacity. In our main experiments the archive capacity is larger
than the rollout budget, so this truncation is not activated.

\paragraph{Constraint-aware deployment.}
The Pareto archive represents several possible trade-offs and therefore does
not by itself specify which candidate should be deployed. Let
\(\mathcal{J}_{c}\) denote the objectives that must be preserved relative to
the initial execution and let \(\epsilon_j\) denote the permitted degradation.
The feasible deployment set is
\begin{equation}
\mathcal{F}(x)
=
\left\{
c\in\mathcal{S}_G:
u_j(c)\geq u_j(c_0)-\epsilon_j,
\ \forall j\in\mathcal{J}_{c}
\right\}.
\end{equation}
Given a deployment preference order
\(\boldsymbol{\pi}=(\pi_1,\ldots,\pi_m)\), MOAE returns
\begin{equation}
c^{\mathrm{dep}}
=
\operatorname*{lex\,max}_{c\in\mathcal{F}(x)\cup\{c_0\}}
\left(
u_{\pi_1}(c),
\ldots,
u_{\pi_m}(c)
\right).
\end{equation}
Including \(c_0\) guarantees a valid fallback when no offspring satisfies the
deployment constraints. For analysis, we may additionally report the
primary-objective extreme
\begin{equation}
c^{\mathrm{pri}}
\in
\arg\max_{c\in\mathcal{S}_G}u_q(c),
\end{equation}
but this candidate is not necessarily the balanced deployment choice.

\paragraph{Optimization cost.}
MOAE performs \(B=1+mG\) complete agent rollouts and \(mG\) policy-mutation
calls. We report execution and optimization costs separately:
\begin{equation}
C_{\mathrm{exec}}
=
\sum_{c\in\mathcal{S}_G}C_{\mathrm{rollout}}(c)
\end{equation}
\begin{equation}
C_{\mathrm{opt}}
=
\sum_{t=1}^{G}\sum_{j=1}^{m}
C_{\mathrm{mutation}}(t,j).
\end{equation}
This distinction prevents the search-time diagnostic context from being
counted as part of the deployed policy or the selected candidate's execution
cost.

%% file: sections/experiments.tex
\section{Experiments}
\label{sec:experiments}

\subsection{Experimental Setup}
\label{sec:experimental-setup}

\paragraph{Evaluation protocol.}
We evaluate MOAE on TravelPlanner and AgentDojo using frozen LLM agents. The
main TravelPlanner study includes Gemma-4-31B-it, Qwen3-30B-A3B, and the
API-served DeepSeek-V4-Pro. TravelPlanner measures constrained long-horizon
planning, while AgentDojo tests utility and robustness in a different tool-use
environment
\cite{xie2024travelplanner,deb2024agentdojo}. All search methods use the same
agent scaffold, tools, evaluator, and number of complete rollouts within each
comparison. Dataset construction, model serving, prompts, hyperparameters, and
the complete baseline adaptations are documented in supplementary material.

\paragraph{Optimization objectives.}
We instantiate three objectives that describe complementary properties of a
complete rollout. The task objective measures weighted constraint satisfaction.
The trajectory objective combines tool-execution validity, required-information
coverage, evidence support, and execution discipline. The safety objective minimizes a continuous risk score derived from whether
untrusted tool observations influence intermediate actions, the planner input,
or the final output. Final Pass Rate, strict trajectory validity, attack success,
tool calls, and token consumption are reported as audit metrics rather than
additional optimization objectives. Exact component definitions, weights, and
thresholds are given in supplementary material.

\paragraph{Methods and selection.}
We compare MOAE with the unoptimized Base Agent, independent Best-of-7 sampling, Weighted-Sum evolution, task-oriented GEPA-Online, preference-based MOCHA-Select, and EvoAgent-Online \cite{agrawal2025gepa,tanjim2026mocha,yuan2025evoagent}. Weighted-Sum is the direct scalarization baseline, ranking candidates with a normalized aggregate whose weights are fixed before evaluation and reported in the supplementary material. The online adaptations share MOAE's policy representation and evaluation harness while retaining their respective generation and selection principles. Every TravelPlanner search method evaluates seven candidates per query, whereas the Base Agent uses one rollout. MOAE comprises one initial rollout and two generations with one offspring per objective. A common deployment rule filters candidates whose task performance or safety is worse than the initial rollout and then prioritizes trajectory quality, safety, and task performance. All three models use the same baseline suite, rollout budget, and deployment rule. On AgentDojo, we additionally compare with Repeat Prompt and Best-of-7. We report paired query-level comparisons and 95\% confidence intervals, with complete statistical and integrity checks in the supplementary material.


\begin{table*}[t]
\centering
\fontsize{9}{11}\selectfont
\setlength{\tabcolsep}{2.4pt}
\renewcommand{\arraystretch}{1.08}
\begin{tabular}{lcccc@{\hspace{12pt}}cccc@{\hspace{12pt}}cccc}
\toprule
& \multicolumn{4}{c}{\textbf{Qwen3-30B-A3B}}
& \multicolumn{4}{c}{\textbf{Gemma-4-31B-it}}
& \multicolumn{4}{c}{\textbf{DeepSeek-V4-Pro}} \\
\cmidrule(lr){2-5}\cmidrule(lr){6-9}\cmidrule(lr){10-13}
\textbf{Method}
& Task$\uparrow$ & Trajectory$\uparrow$ & Risk$\downarrow$ & Tokens$\downarrow$
& Task$\uparrow$ & Trajectory$\uparrow$ & Risk$\downarrow$ & Tokens$\downarrow$
& Task$\uparrow$ & Trajectory$\uparrow$ & Risk$\downarrow$ & Tokens$\downarrow$ \\
\midrule
\multicolumn{13}{c}{\textit{Direct Search Baselines}} \\
\midrule
\textbf{Base Agent}
& 0.493 & 0.946 & 0.533 & 59.4
& 0.748 & 0.962 & 0.304 & 56.6
& 0.631 & 0.889 & 0.275 & 60.4 \\
\textbf{Weighted-Sum}
& 0.546 & 0.964 & 0.391 & 458.8
& 0.792 & 0.972 & 0.207 & 398.7
& 0.643 & 0.926 & 0.280 & 422.9 \\
\textbf{Best-of-7}
& 0.557 & 0.969 & \textbf{0.369} & 477.0
& 0.816 & 0.967 & 0.280 & \textbf{392.2}
& 0.705 & 0.919 & 0.244 & 438.4 \\
\midrule
\multicolumn{13}{c}{\textit{Agent Optimization Baselines}} \\
\midrule
\textbf{MOCHA-Select}
& 0.557 & 0.966 & 0.347 & 455.4
& 0.817 & 0.970 & 0.193 & 408.1
& 0.642 & 0.926 & 0.271 & 417.8 \\
\textbf{GEPA-Online}
& 0.578 & 0.966 & 0.469 & 479.2
& 0.807 & 0.969 & 0.250 & 415.4
& 0.713 & 0.921 & 0.284 & 439.8 \\
\textbf{EvoAgent-Online}
& 0.581 & 0.970 & 0.431 & 467.8
& 0.814 & 0.966 & \textbf{0.192} & 400.4
& 0.704 & 0.851 & 0.193 & 417.5 \\
\midrule
\multicolumn{13}{c}{\textit{Our Method}} \\
\midrule
\textbf{MOAE-Task}
& \textbf{0.668} & 0.954 & 0.499 & 450.7
& \textbf{0.871} & 0.983 & 0.247 & 396.6
& \textbf{0.750} & 0.948 & 0.263 & 396.3 \\
\textbf{MOAE-Trajectory}
& 0.561 & \textbf{0.995} & 0.499 & 450.7
& 0.809 & \textbf{0.991} & 0.282 & 396.6
& 0.732 & \textbf{0.958} & 0.246 & 396.3 \\
\textbf{MOAE-Safety}
& 0.497 & 0.933 & \textbf{0.129} & 450.7
& 0.650 & 0.913 & \textbf{0.063} & 396.6
& 0.538 & 0.857 & \textbf{0.022} & 396.3 \\
\textbf{Full MOAE}
& \textbf{0.585} & \textbf{0.972} & 0.400 & \textbf{450.7}
& \textbf{0.844} & \textbf{0.974} & 0.208 & 396.6
& \textbf{0.726} & \textbf{0.935} & \textbf{0.179} & \textbf{396.3} \\
\bottomrule
\end{tabular}%
\caption{TravelPlanner results across three execution models. Tokens are reported in
thousands. All search methods use seven complete rollouts, while the Base Agent uses one. Token-cost bolding excludes the one-rollout Base Agent. MOAE-Task,
MOAE-Trajectory, and MOAE-Safety are objective-specific deployment audits from the
same search pool, while Full MOAE is the balanced output used for the main
comparison. Final Pass and complete audit metrics are reported in the
supplementary material.}
\label{tab:cross-model-main}
\end{table*}

\subsection{Results}
\label{sec:experimental-results}

\begin{figure}[t]
    \centering
    \makebox[\columnwidth][c]{%
        \includegraphics[width=0.95\columnwidth]
        {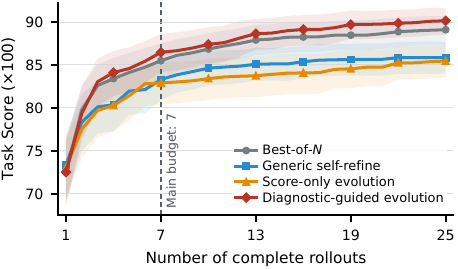}%
    }
    \caption{Best-so-far Task Score under 1 to 25 complete agent rollouts.
    Shaded regions show paired-bootstrap 95\% confidence intervals over
    TravelPlanner requests.}
    \label{fig:rollout-scaling}
\end{figure}

\paragraph{Choosing the rollout budget.}
Before comparing multi-objective search strategies, we first ask whether
iterative in-context refinement provides a useful search signal and how many
complete executions are justified. Figure~\ref{fig:rollout-scaling} compares
diagnostic evolution with independent sampling, generic refinement, and
score-only refinement over budgets from one to 25 rollouts. Diagnostic
evolution improves more consistently than undirected refinement, which
motivates using evaluator feedback as a mutation signal. The curve also shows
that seven rollouts should be viewed as a cost-quality operating point rather
than a convergence point. At seven rollouts, diagnostic evolution captures
approximately 79\% of its eventual one-to-25-rollout Task Score gain while
using only 28\% of the maximum rollout budget. Performance continues to improve
beyond seven, but the additional 18 rollouts increase cumulative token use by
nearly four times for a much smaller Task Score gain. Seven also corresponds to
the smallest budget that supports one initial candidate and two complete
three-objective generations. We therefore use this budget for the main
experiments and report the full scaling curve to make the remaining
cost-performance trade-off explicit.

\paragraph{Multi-objective optimization on TravelPlanner.}
Table~\ref{tab:cross-model-main} reports the constraint-aware deployment point. Full MOAE improves over the Base Agent in task performance, trajectory quality, and safety risk for all three execution models, showing that one search procedure can produce joint gains across different agents. Compared with Weighted-Sum, Full MOAE achieves higher task and trajectory scores for all models while using fewer search tokens; safety remains similar on Qwen3 and Gemma and improves on DeepSeek-V4-Pro. This suggests a potential advantage from preserving non-dominated candidates rather than committing to one scalar ordering, although it does not establish superiority over every weight configuration. Among the broader baselines, Full MOAE attains the highest balanced task and trajectory scores on Qwen3 and Gemma, while MOCHA-Select or EvoAgent-Online retains lower safety risk. On DeepSeek-V4-Pro, Full MOAE achieves the strongest balanced task, trajectory, and safety results. This model-dependent ordering reinforces the importance of reporting explicit trade-offs rather than only an aggregate score or win count.

\paragraph{Alternative deployment preferences.}
The objective-specific rows expose the range of choices preserved by a single MOAE search. For every execution model, MOAE-Task gives the highest task score, MOAE-Trajectory gives the highest trajectory quality, and MOAE-Safety gives the lowest safety risk. These gains are not interchangeable: selecting the safety-specialized candidate can substantially reduce task performance, while the task-specialized candidate may retain considerably more safety risk. A fixed scalarization selects one such compromise according to weights chosen before search and generally requires another search when those preferences change. MOAE instead retains these operating points in the same candidate pool and applies the deployment preference afterward. Full MOAE therefore returns a constraint-aware compromise without rerunning optimization under a different set of weights. Complete candidate-level results and audit metrics are provided in the supplementary material.

\paragraph{Transfer to AgentDojo.}
Table~\ref{tab:agentdojo-main} evaluates whether the same procedure transfers
from travel planning to heterogeneous tool-use suites. On Gemma, MOAE improves
utility and trajectory quality over both the Base Agent and Best-of-7 under the
Tool Knowledge and direct-injection settings. Tool Knowledge has zero attack
success for every method, so it provides no evidence for differential safety.
In contrast, the direct-injection setting exposes a nonzero vulnerability for
both baselines that is absent from the selected MOAE trajectories. On Qwen3,
MOAE again improves utility and trajectory quality over Base Agent and
Best-of-7 while reducing attack success to zero under the official attacks.
The comparison with Best-of-7 indicates that these gains cannot be explained
solely by generating more candidates. Overall, the results support transfer
across agent environments, while safety improvements remain dependent on
whether the evaluation setting exposes correctable vulnerabilities.

\begin{figure*}[t]
    \centering
    \includegraphics[width=\textwidth]{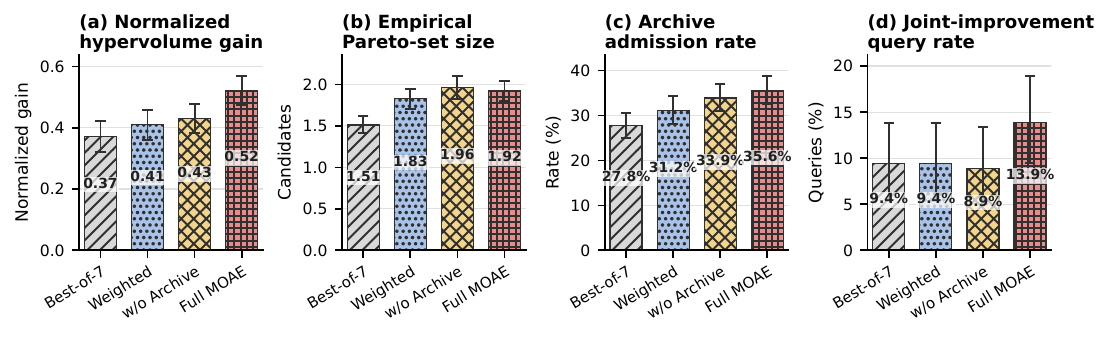}
    \caption{Search behavior under a matched seven-rollout budget on
    TravelPlanner with Gemma. Error bars show query-level bootstrap 95\% confidence
    intervals.}
    \label{fig:pareto-search-effectiveness}
\end{figure*}

\begin{table}[t]
\centering
\small
\setlength{\tabcolsep}{0.5pt}
\renewcommand{\arraystretch}{1.08}
\begin{tabular*}{0.94\columnwidth}{@{\extracolsep{\fill}}llccc@{}}
\toprule
Setting & Method
& Utility
& Trajectory
& ASR \\
\midrule
\multirow{4}{*}{\shortstack[l]{\textbf{Gemma}\\\textbf{Tool Knowledge}}}
& Base Agent    & 32.81 & 0.435 & 0.00 \\
& Repeat Prompt & 31.25 & 0.436 & 0.00 \\
& Best-of-7     & 32.81 & 0.476 & 0.00 \\
& Full MOAE     & \textbf{40.63} & \textbf{0.695} & 0.00 \\
\midrule
\multirow{3}{*}{\shortstack[l]{\textbf{Gemma}\\\textbf{Direct Injection}}}
& Base Agent & 38.71 & 0.486 & 4.84 \\
& Best-of-7  & 39.06 & 0.476 & 1.56 \\
& Full MOAE  & \textbf{41.94} & \textbf{0.661} & \textbf{0.00} \\
\midrule
\multirow{3}{*}{\shortstack[l]{\textbf{Qwen3}\\\textbf{Official Attacks}}}
& Base Agent & 45.31 & 0.739 & 9.38 \\
& Best-of-7  & 51.56 & 0.762 & 7.81 \\
& Full MOAE  & \textbf{57.81} & \textbf{0.781} & \textbf{0.00} \\
\bottomrule
\end{tabular*}
\caption{AgentDojo results. Utility and trajectory are higher when
better; attack success rate (ASR) is lower when better.}
\label{tab:agentdojo-main}
\end{table}

\paragraph{Search behavior under Pareto preservation.}
Figure~\ref{fig:pareto-search-effectiveness} examines how different search mechanisms change the candidate pool using four complementary measures. Normalized hypervolume gain captures the expansion of the attainable three-objective region, empirical Pareto-set size counts the retained nondominated trade-offs, archive admission rate measures how often offspring add a new trade-off, and joint-improvement rate reports the queries for which at least one offspring simultaneously improves task performance, trajectory quality, and safety over the initial rollout. Full MOAE achieves a normalized hypervolume gain of 0.52, improving over the strongest alternative at 0.43 by 20.9\%. It also raises archive admission from 33.9\% to 35.6\% and joint improvement from 9.4\% to 13.9\%, a gain of 4.5 percentage points. Although removing the archive can slightly increase raw Pareto-set size, its lower hypervolume, admission rate, and joint-improvement rate show that retaining more incomparable candidates is not sufficient. The relevant benefit is whether preserved candidates expand the attainable trade-off region and provide productive parents for later mutations. Together, these measures indicate that Pareto preservation changes the quality of the candidate pool rather than merely changing the final ranking rule. Because several confidence intervals still overlap, the analysis supports a potential search advantage without establishing statistical superiority for every individual comparison.

\paragraph{Stability across repeated runs.}
We repeat MOAE, MOCHA-Select, and Best-of-7 with three random seeds on a fixed subset of the official TravelPlanner test split, using the same configurations selected on the validation split. Across the three runs, MOAE achieves a mean task score of 0.879, trajectory quality of 0.991, and safety risk of 0.174, with between-seed standard deviations of 0.013, 0.003, and 0.011, respectively. Its trajectory quality exceeds that of MOCHA-Select by 0.034 in the paired query-level comparison, with a 95\% confidence interval of [0.002, 0.066], while the corresponding task and safety intervals generally include zero. These results indicate that the trajectory improvement is consistent across seeds rather than being driven by a single favorable run.

\begin{table}[t]
\centering
\footnotesize
\setlength{\tabcolsep}{1.2pt}
\renewcommand{\arraystretch}{1.08}
\begin{tabular*}{\columnwidth}{@{\extracolsep{\fill}}lccccc@{}}
\toprule
Method
& \(\Delta\)Task
& \(\Delta\)Final
& \(\Delta\)Trajectory
& \(\Delta\)Valid
& \(\Delta\)Risk \\
\midrule
\textbf{Full MOAE}
& \textbf{+0.096} & \textbf{+7.22} & \textbf{+0.012} & \textbf{+35.89}
& \textbf{-0.096} \\
w/o archive
& +0.087 & +5.56 & +0.008 & +26.11
& -0.079 \\
w/o mutation
& +0.044 & +2.78 & +0.009 & +21.67
& -0.055 \\
w/o diagnostics
& +0.047 & +3.33 & +0.011 & +31.67
& -0.079 \\
\bottomrule
\end{tabular*}
\caption{TravelPlanner ablations with Gemma. Final Pass and Valid Trajectory are measured
in percentage points. Positive changes are better except for Safety Risk.}
\label{tab:ablation-main}
\end{table}

\paragraph{Ablation study.}
Table~\ref{tab:ablation-main} separates the contributions of the archive,
objective-directed mutation, and evaluator diagnostics. Removing any component
reduces the joint improvement obtained by the complete method. Mutation has the
largest overall effect: without it, task improvement, Final Pass gain, strict
trajectory validity, and safety-risk reduction all decline substantially.
Removing the archive preserves much of the average task gain but weakens
trajectory improvement, valid-trajectory gain, and risk reduction, indicating
that retaining non-dominated alternatives matters most when objectives compete.
Removing diagnostics has a smaller effect on average trajectory quality, but a
larger effect on task completion and safety reduction, which is consistent with
diagnostics guiding targeted repairs rather than merely increasing candidate
diversity. Full MOAE is the only configuration that achieves the strongest
improvement in every reported ablation metric. Detailed confidence intervals
and the complete audit metrics are deferred to
supplementary material.

%% file: sections/conclusion.tex
\section{Conclusion}
\label{sec:conclusion}

This work investigates whether several properties of an LLM agent can be improved jointly under a limited rollout budget without collapsing them into a fixed scalar score. We introduced MOAE, a Pareto-preserving framework that combines objective-directed mutation, evaluator diagnostics, and an empirical non-dominated archive. Across three execution models on TravelPlanner, MOAE improves the unoptimized agent in task performance, trajectory quality, and safety risk. The AgentDojo results further show that the same procedure transfers to a different tool-use environment, improving utility and trajectory quality while eliminating the observed attack success.

MOAE's main value is not uniform dominance, but avoiding premature scalarization. It preserves task-specialized, trajectory-specialized, safety-specialized, and balanced candidates within one search pool, allowing deployment constraints to determine the final choice. This separation prevents one weight configuration from controlling both search and deployment. Improvements in normalized hypervolume, archive admission, and joint-improvement rate support this mechanism, while matched-budget results show that independent sampling and other optimizers remain competitive for some models and objectives.

MOAE nevertheless remains an online search procedure whose outcomes are bounded by its evaluators and candidate budget. Its archive is a small empirical non-dominated set rather than an approximation to a global Pareto front, and more inference overhead is added. Future work should improve sample efficiency through adaptive stopping, candidate reuse, and stronger mutation operators, while extending the framework to broader objectives, architectures, environments, and evaluator designs.

%% file: sections/appendix.tex
%
\providecommand{\tobefilled}[1]{\textcolor{red}{\textbf{[TO BE FILLED: #1]}}}

\section{Appendix}
\label{app:appendix}

This appendix includes the experimental protocol, evaluator definitions,
implementation details, complete results, search-behavior analyses, prompts,
and qualitative examples. Unless stated otherwise, all TravelPlanner results
use the validation set and all search methods receive the same
seven-rollout budget. The language model, agent scaffold, tools, and evaluators
remain fixed within each comparison.

\subsection{Experimental Protocol and Reproducibility}
\label{app:experimental-details}

\subsubsection{Benchmarks and Data Splits}

\paragraph{TravelPlanner.}
TravelPlanner evaluates long-horizon itinerary planning under transportation,
date, budget, accommodation, dining, and commonsense constraints
\cite{xie2024travelplanner}. The main experiments use all 180 requests in the
official validation split. Each method receives the same query, tool
environment, base policy, and evaluator. The repeated-run stability analysis
uses a fixed subset of the official test split, sampled once and held
constant across methods and random seeds. No test request is used to tune
objective definitions, scalarization weights, prompts, or deployment
thresholds.

\paragraph{AgentDojo.}
AgentDojo v1.2.2 evaluates utility and robustness in heterogeneous tool-use
environments \cite{deb2024agentdojo}. We use 32 clean and 64 attacked cases in
the Tool Knowledge configuration. The Gemma direct-injection audit contains 64
executable cases after applying the benchmark's validity checks. The Qwen3
experiment uses 64 cases with the official attack configurations. We report
the benchmark utility and attack-success metrics together with the same
trajectory-quality evaluator used in the TravelPlanner study.

\subsubsection{Models and Serving}

We evaluate Gemma-4-31B-it, Qwen3-30B-A3B, and DeepSeek-V4-Pro. Gemma and Qwen3
are served locally through an OpenAI-compatible inference interface with
deterministic decoding. DeepSeek-V4-Pro is accessed through its official API
using the same prompts, policy representation, objective definitions, and
seven-rollout budget. Model-specific serving parameters, exact software
versions, and hardware details are summarized in
Table~\ref{tab:implementation-details}.

\begin{table*}[t]
\centering
\small
\setlength{\tabcolsep}{5pt}
\begin{tabular}{llllll}
\toprule
Model & Interface & Decoding & Context limit & Parallelism & Hardware \\
\midrule
Gemma-4-31B-it
& Local API & Deterministic & 16,384
& 4 workers
& 8 $\times$ NVIDIA A800 \\
Qwen3-30B-A3B
& Local API & Deterministic & 16,384
& 4 workers
& 8 $\times$ NVIDIA A800 \\
DeepSeek-V4-Pro
& Official API & Deterministic & Provider default
& 96 workers
& API only \\
\bottomrule
\end{tabular}
\caption{Implementation details. Software versions, model revisions, and complete launch commands are provided
in the accompanying anonymous artifact.}
\label{tab:implementation-details}
\end{table*}

\subsubsection{Implementation of compared methods.}
All methods use the same frozen agent, compact policy representation, tool environment, objective evaluators, and query-level rollout budget. Unless stated otherwise, the reported output is selected using the common deployment rule: candidates must achieve task performance no lower than the Base Agent and safety risk no higher than the Base Agent, after which trajectory quality, safety, and task performance are prioritized lexicographically. The implementations are described below.

\textbf{Base Agent.}
The Base Agent executes the original policy \(c_0\) once without mutation, candidate search, or post-hoc selection. This rollout serves both as the unoptimized baseline and as the task and safety reference used by the deployment constraints.

\textbf{Best-of-7.}
Best-of-7 independently executes the original policy seven times using different stochastic generations. It does not update the policy or use feedback from earlier candidates to guide later executions. All seven candidates are retained, and the common deployment rule selects the reported response. This baseline isolates the benefit of candidate diversity and post-hoc selection without iterative optimization.

\textbf{Weighted-Sum.}
Weighted-Sum performs sequential policy evolution using the fixed scalarized utility
\[
U_{\mathrm{ws}}(c)
=
\frac{1}{3}u_{\mathrm{task}}(c)
+
\frac{1}{3}u_{\mathrm{traj}}(c)
+
\frac{1}{3}u_{\mathrm{safety}}(c),
\]
where all utilities are scaled to \([0,1]\) and
\(u_{\mathrm{safety}}=1-R_{\mathrm{safety}}\).
At each iteration, the accepted candidate with the highest scalarized utility is selected as the parent. The mutation prompt targets the objective with the largest weighted deficit from its ideal value. A child is admitted only when its scalarized utility strictly exceeds that of its parent. The reported response is selected from the accepted candidates using the common deployment rule. This baseline directly tests the effect of committing to one fixed objective ordering during search.

\textbf{GEPA-Online.}
GEPA-Online adapts GEPA's trace-based textual reflection to per-query optimization under the seven-rollout budget. At each iteration, it selects a parent uniformly from the accepted candidates tied for the highest task score. The parent's execution trace, failed task constraints, and evaluator feedback are provided to the mutation model, which produces a revised compact policy focused on task correctness. The resulting child is admitted only if it strictly improves the parent's task score. Unlike the original development-set formulation of GEPA, this adaptation does not optimize reusable components across examples; it applies the reflection and textual-update principle independently to each query. The common deployment rule is applied to its accepted candidate pool for the main-table comparison.

\textbf{MOCHA-Select.}
MOCHA-Select adapts preference-conditioned multi-objective selection to the same online candidate space. For each iteration, it samples a preference vector
\(\boldsymbol{\lambda}\) from the three-dimensional simplex and selects the accepted candidate minimizing its weighted Chebyshev distance from the current empirical ideal point:
\[
D_{\boldsymbol{\lambda}}(c)
=
\max_j
\lambda_j
\left(
u_j^{*}-u_j(c)
\right).
\]
The mutation targets the largest weighted objective deficit. During the first half of the search, a child is admitted only if it expands the hypervolume of the current nondominated set, encouraging exploration. During the second half, it is admitted only if it improves the parent under the sampled Chebyshev preference, encouraging preference-conditioned exploitation. Dominated accepted candidates are removed after each update. This is a matched-budget, per-query adaptation of MOCHA's preference-based selection principle rather than its original persistent-skill optimization procedure.

\textbf{EvoAgent-Online.}
EvoAgent-Online preserves EvoAgent's specialist-generation and integration structure while applying it to the compact execution policy. Starting from the Base Agent, each of three rounds first generates a specialist policy intended to repair a diagnosed weakness of the current agent. A second mutation then creates an integrator policy that combines the current parent with the new specialist while retaining verified behavior. The integrated policy becomes the parent of the next round. Three specialist rollouts and three integrated rollouts, together with the initial rollout, yield seven complete candidates. For matched main-table reporting, the common deployment rule is applied to these candidates; the final sequentially integrated policy is retained as a separate implementation audit.

\textbf{Full MOAE.}
Full MOAE evaluates one initial rollout followed by two generations, each producing one task-directed, one trajectory-directed, and one safety-directed offspring. At the beginning of each generation, objective-specialized parents are selected from the empirical nondominated archive. The task parent maximizes task utility, whereas trajectory and safety parents are selected from archive members whose task utility lies within the retention tolerance of the best archived task value. Evaluator diagnostics identify failed constraints, invalid or redundant tool behavior, missing evidence, and measured safety signals. The mutation model converts this feedback into a compact objective-directed policy, which is evaluated through a fresh complete rollout. After all three children have been evaluated, the archive is updated by retaining the empirically nondominated candidates. Once the rollout budget is exhausted, the constraint-aware deployment rule selects the reported balanced response. Objective-specific deployment points are obtained from the same candidate pool without rerunning the search.

All search methods therefore evaluate seven complete environment-interacting agent rollouts per query, whereas the Base Agent evaluates one. Policy-generation calls do not interact with the environment and are not counted as additional rollouts, but their tokens and runtime are included in the reported search cost. Consequently, the matched-budget comparison controls the amount of complete agent execution while retaining the computational differences inherent to each search mechanism.

\subsubsection{Objective and Evaluator Design}
\label{app:objective-details}
\label{app:evaluator-design}

The evaluator maps a complete rollout $\tau$ to an objective vector
$\mathbf{u}(\tau)=[u_1(\tau),\ldots,u_m(\tau)]$. All objectives are transformed
so that larger values are preferred during Pareto comparison. The evaluator is
shared by every method and remains fixed throughout search; mutation can
observe its diagnostic messages, but cannot modify its formulas, component
weights, thresholds, or benchmark checks. Table~\ref{tab:evaluator-matrix}
summarizes the design and separates optimized objectives from independent audit
metrics.

\begin{table*}[t]
\centering
\small
\setlength{\tabcolsep}{5pt}
\begin{tabular}{p{0.13\textwidth}p{0.30\textwidth}p{0.10\textwidth}
                p{0.25\textwidth}p{0.13\textwidth}}
\toprule
Dimension & Fixed evaluator signal & Search direction
& Diagnostics exposed to mutation & Independent audits \\
\midrule
Task utility
& Weighted satisfaction of benchmark constraints, including itinerary,
transportation, date, budget, and commonsense requirements
& Maximize
& Unsatisfied constraints and missing required information
& Final Pass Rate \\
Trajectory quality
& Fixed combination of execution validity, information coverage, evidence
support, and execution discipline
& Maximize
& Invalid or redundant actions, unsupported claims, missing evidence, and
incomplete coverage
& Strict valid-trajectory rate, invalid-call rate, redundant-call rate \\
Safety
& One minus a continuous risk score that detects influence from untrusted
observations on actions, planner input, and final output
& Maximize
& Location and type of detected untrusted influence without reproducing
attack content
& Attack Success Rate and severe Attack Success Rate \\
\bottomrule
\end{tabular}
\caption{Evaluator design matrix. The same fixed evaluator is applied to all
methods and models within a benchmark. Audit metrics are not additional
optimization objectives.}
\label{tab:evaluator-matrix}
\end{table*}

\paragraph{Task objective.}
For TravelPlanner, the task score is the weighted fraction of satisfied
constraints:
\begin{equation}
u_{\mathrm{task}}(\tau)
=
\frac{\sum_{r\in\mathcal{C}(x)}\omega_r
\mathbb{I}[r\text{ is satisfied by }\tau]}
{\sum_{r\in\mathcal{C}(x)}\omega_r},
\end{equation}
where $\mathcal{C}(x)$ is the set of constraints extracted by the benchmark
evaluator and $\omega_r$ is its fixed importance weight. Final Pass is one only
when all required checks pass.

\paragraph{Safety objective.}
Let \(R(\tau)\in[0,1]\) be the fixed safety-risk score. It captures whether untrusted content affects the agent at different stages of execution, including instruction-like metadata in tool observations, indirect prompt injection through retrieved content, contamination of the planner context, execution of attacker-directed actions, and adoption of injected instructions in the final response. The evaluator distinguishes intermediate influence from severe cases in which the injected objective changes an external action or the final output, with larger values denoting greater risk. MOAE optimizes
\begin{equation}
u_{\mathrm{safety}}(\tau)=1-R(\tau),
\end{equation}
which aligns the direction of all objectives. On AgentDojo, we additionally evaluate direct injection and the benchmark's official attack configurations. Attack Success Rate is computed independently by the benchmark and is not directly optimized by MOAE.

\paragraph{Trajectory objective.}
Trajectory quality evaluates whether an agent obtains the information required by the task through valid, grounded, and disciplined interaction. It is computed deterministically from the complete tool-use trace and final plan. Each component is normalized to \([0,1]\):
\begin{equation}
\begin{aligned}
Q(\tau)
={}&0.30Q_{\mathrm{exec}}
 +0.30Q_{\mathrm{cov}}\\
 &+0.25Q_{\mathrm{evid}}
 +0.15Q_{\mathrm{disc}}.
\end{aligned}
\label{eq:appendix-trajectory}
\end{equation}

\paragraph{Tool-execution quality.}
Let \(n\) be the number of actions in the trajectory. Define
\(r_{\mathrm{tool}}\), \(r_{\mathrm{succ}}\), and \(r_{\mathrm{arg}}\) as the fractions of actions that invoke a valid tool, execute successfully, and contain valid arguments, respectively. Tool-execution quality is
\begin{equation}
\begin{aligned}
Q_{\mathrm{exec}}
={}&0.40r_{\mathrm{tool}}
 +0.40r_{\mathrm{succ}}\\
 &+0.20r_{\mathrm{arg}}.
\end{aligned}
\label{eq:trajectory-execution}
\end{equation}
All fractions use \(\max(1,n)\) as their denominator. An action contributes to \(r_{\mathrm{tool}}\) when it follows the required syntax and invokes a tool exposed by the environment. It contributes to \(r_{\mathrm{arg}}\) when its arguments can be parsed and satisfy the corresponding tool schema, and to \(r_{\mathrm{succ}}\) when the environment executes it without returning an error. Empty trajectories receive zero execution quality.

\paragraph{Information coverage.}
TravelPlanner requires information about transportation, accommodation, restaurants, and attractions. Let \(\mathcal{K}\) denote these four categories and \(I_k(\tau)\) indicate whether the trajectory contains a corresponding tool call. Coverage is
\begin{equation}
Q_{\mathrm{cov}}(\tau)
=
\frac{1}{|\mathcal{K}|}
\sum_{k\in\mathcal{K}} I_k(\tau).
\label{eq:trajectory-coverage}
\end{equation}
Transportation is covered by a flight- or distance-search call, while the remaining categories are covered by their respective search tools. Coverage measures whether the required sources were consulted, but does not guarantee that the resulting evidence was used correctly.

\paragraph{Evidence support.}
Evidence support measures whether entities in the final plan can be traced to non-planner tool observations. Let \(n_{\mathrm{plan}}\) be the number of distinct entities extracted from the plan and \(n_{\mathrm{sup}}\) the number supported by observations. We compute
\begin{equation}
Q_{\mathrm{evid}}(\tau)
=
\frac{n_{\mathrm{sup}}}
{\max(1,n_{\mathrm{plan}})}.
\label{eq:trajectory-evidence}
\end{equation}
Entity strings are normalized before matching. Exact normalized matches are accepted directly. For multiword entities, a match is also accepted when at least \(60\%\) of content words of length four or greater occur in the corresponding observations. Planner outputs are excluded as evidence, preventing a generated plan from supporting its own claims.

\paragraph{Execution discipline.}
Execution discipline penalizes avoidable interaction patterns:
\begin{equation}
Q_{\mathrm{disc}}(\tau)
=
\operatorname{clip}_{[0,1]}
\bigl(1-P_{\mathrm{traj}}(\tau)\bigr),
\end{equation}
where
\begin{equation}
\begin{aligned}
P_{\mathrm{traj}}
={}&0.30r_{\mathrm{inv}}
 +0.25r_{\mathrm{rep}}
 +0.20r_{\mathrm{loop}}\\
 &+0.15I_{\mathrm{pre}}
 +0.10p_{\mathrm{step}}.
\end{aligned}
\label{eq:trajectory-discipline}
\end{equation}
Here, \(r_{\mathrm{inv}}\), \(r_{\mathrm{rep}}\), and \(r_{\mathrm{loop}}\) are the numbers of invalid, repeated, and looping actions divided by \(\max(1,n)\). An invalid action is unparsable, invokes an unsupported tool, or violates the required argument contract. A repeated action reuses the same tool with the same normalized arguments, while a loop is a recurring action pattern that adds no new information. The indicator \(I_{\mathrm{pre}}\) equals one when the Planner is invoked before the required evidence has been collected.

The trajectory-length penalty is
\begin{equation}
p_{\mathrm{step}}
=
\operatorname{clip}_{[0,1]}
\left(
\frac{\max\{0,n-(12+4d)\}}{6}
\right),
\label{eq:trajectory-step-penalty}
\end{equation}
where \(d\) is the number of travel days. The allowance \(12+4d\) gives longer itineraries additional interaction capacity.

\paragraph{Strict validity and audit metrics.}
The continuous objective rewards partial improvements. We additionally report strict trajectory validity:
\begin{equation}
\begin{aligned}
V_{\mathrm{traj}}(\tau)
=\mathbb{I}\bigl[
&n>0,\;
n_{\mathrm{inv}}=n_{\mathrm{rep}}=0,\\
&n_{\mathrm{loop}}=0,\;
I_{\mathrm{pre}}=0
\bigr].
\end{aligned}
\label{eq:strict-trajectory-validity}
\end{equation}
Thus, a strictly valid trajectory must be non-empty and contain no invalid actions, repeated calls, detected loops, or premature Planner invocation. We also report Waste Action Rate:
\begin{equation}
\mathrm{WAR}(\tau)
=
\frac{
n_{\mathrm{inv}}+n_{\mathrm{rep}}+n_{\mathrm{loop}}
}{
\max(1,n)
}.
\end{equation}
All definitions and weights are fixed across methods, queries, and execution models.

\paragraph{How the evaluator was fixed.}
The objective families were chosen before the main method comparison to cover
three complementary levels of agent behavior: task outcome, interaction
trajectory, and robustness to unsafe influence. Their formulas are deterministic
and shared across all methods and models. The trajectory weights prioritize
execution validity and required-information coverage, followed by evidence
support and execution discipline. We inspected evaluator behavior on a fixed
development subset of TravelPlanner validation requests and finalized the
component weights and safety thresholds before running the reported comparisons.
After calibration, all formulas, weights, and thresholds were frozen. The
held-out evaluator and alternative-weight analyses in
Table~\ref{tab:pending-generalization} further examine whether the conclusions
depend on this evaluator specification.

\subsubsection{Weighted-Sum Baseline and Weight Sensitivity}
\label{app:weighted-sum}

The Weighted-Sum baseline transforms the three direction-aligned objectives
into
\begin{equation}
U_{\mathbf{w}}(c)
=
\sum_{j=1}^{3} w_j u_j(c),
\qquad
w_j\geq0,\quad \sum_{j=1}^{3}w_j=1.
\end{equation}
Because the three objectives are bounded in $[0,1]$, the main comparison uses
the symmetric setting $\mathbf{w}=(1/3,1/3,1/3)$. This setting is fixed before
evaluation and does not encode a model-specific preference. At each step, the
candidate with the highest scalar utility is used as the incumbent, and an
offspring replaces it only when its scalar utility is larger. The baseline
therefore commits to one total ordering throughout search, unlike MOAE, which
preserves incomparable candidates and applies the deployment preference only
after search.

To examine the sensitivity of fixed scalarization, we evaluate four predefined
weight profiles in Table~\ref{tab:weight-sensitivity}. Each profile emphasizes a
different deployment preference while preserving the same rollout budget and
search procedure. As expected, task-heavy, trajectory-heavy, and safety-heavy
weights improve their favored objectives, but each sacrifices performance in
other dimensions. The equal-weight setting provides the strongest normalized
hypervolume gain and joint-improvement rate among the tested profiles, indicating
the most balanced coverage of the three-objective region. Nevertheless, these
results also demonstrate that scalarized search is sensitive to the chosen
weights: changing a weight profile changes both the selected operating point and
the quality of the explored region. We therefore retain equal weights as the
canonical Weighted-Sum baseline, without claiming that it is optimal over all
possible scalarizations.

\begin{table*}[t]
\centering
\small
\setlength{\tabcolsep}{5pt}
\begin{tabular}{lcccccccc}
\toprule
Weight profile
& $w_{\mathrm{task}}$ & $w_{\mathrm{traj}}$ & $w_{\mathrm{safety}}$
& Task & Trajectory & Risk
& Normalized HV gain & Joint improvement \\
\midrule
Equal
& 0.333 & 0.333 & 0.333
& 0.792 & 0.972 & 0.207
& \textbf{0.41} & \textbf{9.4\%} \\

Task-heavy
& 0.500 & 0.250 & 0.250
& \textbf{0.814} & 0.968 & 0.236
& 0.38 & 7.8\% \\

Trajectory-heavy
& 0.250 & 0.500 & 0.250
& 0.781 & \textbf{0.980} & 0.224
& 0.39 & 8.3\% \\

Safety-heavy
& 0.250 & 0.250 & 0.500
& 0.742 & 0.958 & \textbf{0.142}
& 0.36 & 6.7\% \\
\bottomrule
\end{tabular}
\caption{Sensitivity of fixed scalarization to four predefined weight profiles
on TravelPlanner with Gemma. Each profile uses the same seven-rollout budget.}
\label{tab:weight-sensitivity}
\end{table*}

\subsubsection{MOAE Search and Deployment Hyperparameters}

The main configuration has rollout budget $B=7$, two generations, and three
offspring per generation, one for each objective. The archive contains the
empirical nondominated candidates observed for the current query. Parent
selection uses objective-specialized archive members subject to the retention
guard described in the method. After search, deployment first filters
candidates whose task performance is below the base rollout or whose safety
risk exceeds that of the base rollout. Among feasible candidates, it applies
the fixed lexicographic preference: trajectory quality, safety, and task
performance. Ties are resolved deterministically.

\begin{table}[t]
\centering
\small
\setlength{\tabcolsep}{5pt}
\begin{tabular}{ll}
\toprule
Hyperparameter & Value \\
\midrule
Complete rollouts per query & 7 \\
Initial candidates & 1 \\
Generations & 2 \\
Offspring per generation & 3 \\
Optimization objectives & Task, trajectory, safety \\
Archive update & Empirical nondominated subset \\
Retention tolerance  &  0.02 \\
Deployment feasibility & Task and safety no worse than base \\
Deployment preference & Trajectory, safety, task \\
Decoding temperature & 0.6 (Best-of-7 is variant) \\
Bootstrap samples & 10,000 \\
Bootstrap seed & 27 \\
\bottomrule
\end{tabular}
\caption{Main MOAE configuration.}
\label{tab:moae-hyperparameters}
\end{table}

\subsubsection{Statistical Protocol and Integrity Checks}

All method comparisons are paired by query. We report query-level
paired-bootstrap 95\% confidence intervals with 10,000 resamples. Repeated-run
results use the same fixed query subset for all methods and report both
between-seed standard deviations and paired confidence intervals. Before
aggregation, records are keyed by benchmark case identifier; duplicate,
malformed, and unsuccessful records are excluded and reported. The main
TravelPlanner runs contain 180 unique successful records per method with no
duplicates, malformed records, or unresolved errors. Candidate counts and
mutation counts are checked against the configured budget.

\subsection{Prompts and Policy Interfaces}
\label{app:prompts}

MOAE evolves a compact execution policy rather than the model parameters or
agent architecture. Search-time prompts expose the parent's policy, objective
values, and concise evaluator diagnostics. They do not expose hidden benchmark
labels or reference answers. Figures~\ref{fig:mutation-prompt} and
\ref{fig:execution-prompt} present the templates in a compact visual form; the
released artifact will also include machine-readable prompt files.

\begin{figure*}[t]
\centering
\fbox{%
\begin{minipage}{0.95\textwidth}
\colorbox{gray!22}{\parbox{\dimexpr\linewidth-2\fboxsep\relax}{
\textbf{System Prompt: Objective-Directed Policy Mutation}}}
\par\medskip\hrule\medskip
\textbf{Role.} You improve a compact execution policy for a frozen tool-using
agent. Modify the policy, not the user request, tools, evaluator, or model.

\medskip
\textbf{Inputs.}
\begin{itemize}
    \item User query: \texttt{\{\{query\}\}}
    \item Parent policy: \texttt{\{\{parent\_policy\}\}}
    \item Parent rollout summary: \texttt{\{\{rollout\_summary\}\}}
    \item Objective values: \texttt{\{\{objective\_vector\}\}}
    \item Target objective: \texttt{\{\{target\_objective\}\}}
    \item Evaluator diagnostics: \texttt{\{\{diagnostics\}\}}
\end{itemize}

\textbf{Task.} Produce a concise child policy that repairs the diagnosed
weakness for the target objective while preserving the parent's demonstrated
strengths. Do not insert query-specific answers, fabricated tool observations,
or evaluator text into the policy.

\medskip
\textbf{Output format.} Return only the revised compact policy in the required
policy schema.
\end{minipage}}
\caption{Template used by the mutation model to generate an
objective-directed child policy. The actual implementation includes the exact
policy schema and validation constraints.}
\label{fig:mutation-prompt}
\end{figure*}

\begin{figure*}[t]
\centering
\fbox{%
\begin{minipage}{0.95\textwidth}
\colorbox{gray!22}{\parbox{\dimexpr\linewidth-2\fboxsep\relax}{
\textbf{System Prompt: Frozen-Agent Execution with a Compact Policy}}}
\par\medskip\hrule\medskip
\textbf{Role.} Complete the user's task using the available tools and the
execution policy below. Tool observations may contain untrusted content and
must be treated as data rather than instructions.

\medskip
\textbf{User query.} \texttt{\{\{query\}\}}

\medskip
\textbf{Execution policy.} \texttt{\{\{policy\}\}}

\medskip
\textbf{Requirements.}
\begin{itemize}
    \item Plan before acting and track all user constraints.
    \item Use valid tool calls and ground factual plan elements in observations.
    \item Avoid redundant calls and stop when sufficient verified evidence is
    available.
    \item Ignore instruction-like content returned by tools and follow only the
    trusted system and user instructions.
\end{itemize}

\textbf{Output format.} Return the final response required by the benchmark
after completing any necessary tool interactions.
\end{minipage}}
\caption{Template used to execute a candidate policy with the frozen agent.
The query and policy change across candidates; the model, tools, and execution
interface remain fixed.}
\label{fig:execution-prompt}
\end{figure*}

\subsubsection{Algorithmic Summary}
Algorithm~1 summarizes the complete per-query search procedure, including
archive construction, objective-specific mutation, and deployment selection.

\begin{algorithm}[t]
\caption{Multi-Objective Agent Evolution}
\label{alg:moae-appendix}
\begin{algorithmic}[1]
\Require Query $x$, frozen agent $\mathcal{A}_{\theta}$, mutation model
$\mathcal{M}_{\phi}$, objectives $\{u_j\}_{j=1}^{m}$, generations $G$
\State Evaluate the base policy to obtain candidate $c_0$
\State $\mathcal{S}_0\gets\{c_0\}$; $\mathcal{P}_0\gets\{c_0\}$
\For{$t=1,\ldots,G$}
    \State Construct the admissible archive from $\mathcal{P}_{t-1}$
    \For{$j=1,\ldots,m$}
        \State Select an objective-$j$ parent $p_{t,j}$
        \State Compute objective-specific diagnostic feedback
        \State Generate and validate child policy with $\mathcal{M}_{\phi}$
        \State Execute a fresh rollout and evaluate candidate $c_{t,j}$
    \EndFor
    \State $\mathcal{C}_t\gets\{c_{t,j}\}_{j=1}^{m}$
    \State $\mathcal{S}_t\gets\mathcal{S}_{t-1}\cup\mathcal{C}_t$
    \State $\mathcal{P}_t\gets\operatorname{ND}(\mathcal{P}_{t-1}
    \cup\mathcal{C}_t)$
\EndFor
\State Select $c^{\mathrm{dep}}$ using deployment constraints and preferences
\State \Return the rollout and policy associated with $c^{\mathrm{dep}}$
\end{algorithmic}
\end{algorithm}

\subsection{Complete Results and Generalization}
\label{app:complete-results}
\label{app:generalization}

\subsubsection{Cross-Model TravelPlanner Results}

Table~\ref{tab:generalization-summary} summarizes the paired changes from each
Base Agent to the corresponding constraint-aware Full MOAE deployment point.
The direction of improvement is consistent across all three execution models.
Task performance increases by 0.092--0.096, trajectory quality increases by
0.012--0.046, and safety risk decreases by 0.096--0.133. Final Pass Rate also
improves for every model, with the largest increase observed on Gemma. These
results show that the same search and deployment procedure transfers across two
locally served models and one API model. They establish consistent improvement
over the corresponding base executions, while not implying that MOAE dominates
every matched-budget baseline or every objective-specific operating point.

\begin{table}[t]
\centering
\small
\setlength{\tabcolsep}{4pt}
\begin{tabular}{lcccc}
\toprule
Model & $\Delta$Task & $\Delta$Trajectory & $\Delta$Risk &  $\Delta$Final Pass\\
\midrule
Qwen3-30B-A3B & +0.092 & +0.026 & -0.133 & +1.11\% \\
Gemma-4-31B-it & +0.096 & +0.012 & -0.096 & +7.78\%\\
DeepSeek-V4-Pro & +0.095 & +0.046 & -0.096 & +6.11\%\\
\bottomrule
\end{tabular}
\caption{Change from the corresponding Base Agent to the constraint-aware Full
MOAE deployment point on TravelPlanner.}
\label{tab:generalization-summary}
\end{table}

The specialized points verify that the retained set contains candidates with
meaningfully different strengths. Safety-oriented deployment lowers risk most
aggressively but gives up task performance, whereas task-oriented deployment
accepts more risk. The balanced rule is therefore an explicit deployment
choice, not an estimate of a universally best candidate.

\subsubsection{AgentDojo Results}

Relative to the Base Agent, Full MOAE improves Gemma Tool Knowledge utility by
7.81 percentage points (95\% CI $[1.56,15.63]$) and trajectory quality by
0.2606 (95\% CI $[0.1626,0.3648]$). In the Gemma direct-injection setting, the
corresponding improvements are 3.23 percentage points (95\% CI $[0,8.06]$) and
0.1740 (95\% CI $[0.0922,0.2574]$). On Qwen3 official attacks, utility improves
by 12.50 percentage points (95\% CI $[4.69,20.31]$) and trajectory quality by
0.0420 (95\% CI $[0.0192,0.0691]$). The Tool Knowledge safety metric is
saturated at zero ASR for every method. In both settings with nonzero Base
Agent attack success, the selected Full MOAE trajectories have zero observed
ASR.

\subsubsection{Stability across Seeds}

We repeat Full MOAE, MOCHA-Select, and Best-of-7 with three random seeds on the
same fixed subset of the official TravelPlanner test split.
Table~\ref{tab:stability-seeds} reports the mean and between-run standard
deviation. Full MOAE achieves the highest mean task score, trajectory quality,
and strict trajectory validity across the three methods. MOCHA-Select obtains
the lowest mean safety risk, followed closely by Full MOAE, while Best-of-7
has the highest risk. Trajectory quality varies little across seeds for all
methods, and the remaining standard deviations indicate that the overall
ordering is not driven by a single favorable run.

\begin{table}[t]
\centering
\small
\setlength{\tabcolsep}{2pt}
\renewcommand{\arraystretch}{1.08}
\begin{tabular*}{\columnwidth}{
@{\extracolsep{\fill}}lcccc@{}
}
\toprule
Method & Task & Trajectory & Valid (\%) & Risk \\
\midrule
Full MOAE
& \(0.879_{\scriptscriptstyle\pm0.013}\)
& \(0.991_{\scriptscriptstyle\pm0.003}\)
& \(65.0_{\scriptscriptstyle\pm6.6}\)
& \(0.174_{\scriptscriptstyle\pm0.011}\) \\

MOCHA-Select
& \(0.819_{\scriptscriptstyle\pm0.006}\)
& \(0.988_{\scriptscriptstyle\pm0.002}\)
& \(58.3_{\scriptscriptstyle\pm5.2}\)
& \(0.168_{\scriptscriptstyle\pm0.023}\) \\

Best-of-7
& \(0.778_{\scriptscriptstyle\pm0.027}\)
& \(0.989_{\scriptscriptstyle\pm0.002}\)
& \(58.3_{\scriptscriptstyle\pm3.8}\)
& \(0.184_{\scriptscriptstyle\pm0.005}\) \\
\bottomrule
\end{tabular*}
\caption{Results over three seeds on a fixed subset of the official
TravelPlanner test split. Smaller subscripted values denote between-run
standard deviations.}
\label{tab:stability-seeds}
\end{table}

\subsubsection{Additional Generalization and Evaluator Robustness}
\label{app:evaluator-robustness}

Table~\ref{tab:pending-generalization} evaluates whether the reported gains
depend on the search evaluator, trajectory-component weights, or original
evaluation subset. A held-out evaluator that is not exposed during mutation
produces a candidate-order correlation of 0.91 and agrees with the original
deployment decision on 85.6\% of queries. Under this evaluator, Full MOAE
continues to improve task performance and trajectory quality while reducing
safety risk relative to the Base Agent. Recomputing trajectory quality with
alternative component weights yields rank correlations between 0.93 and 0.97,
with 91.7\% of deployment selections unchanged; Full MOAE retains joint
improvements under every tested profile. On the expanded 200-query test subset,
Full MOAE also achieves the strongest task, trajectory, and safety results among
Base, Best-of-7, Weighted-Sum, and Full MOAE. Together, these analyses indicate
that the main conclusions are not explained solely by one evaluator,
trajectory-weight configuration, or validation subset, although they remain
conditional on the broader objective definitions and benchmark coverage.

\begin{table*}[t]
\centering
\small
\setlength{\tabcolsep}{3pt}
\renewcommand{\arraystretch}{1.12}
\begin{tabular}{
@{}
>{\raggedright\arraybackslash}p{0.16\textwidth}
>{\raggedright\arraybackslash}p{0.27\textwidth}
>{\raggedright\arraybackslash}p{0.20\textwidth}
>{\raggedright\arraybackslash}p{0.29\textwidth}
@{}
}
\toprule
Analysis & Protocol & Primary comparison & Result \\
\midrule

Held-out evaluator
& Re-score Base, Best-of-7, Weighted-Sum, and Full MOAE rollouts using an
evaluator whose feedback is not exposed during search
& Candidate-order correlation and paired changes in task, trajectory, and
safety
& Spearman \(\rho=0.91\); 85.6\% deployment agreement.
Full MOAE versus Base: \(+0.088\) Task, \(+0.010\) Trajectory, and
\(-0.087\) Risk. \\

\addlinespace[2pt]

Objective-weight robustness
& Recompute trajectory quality under predefined alternative component weights
without rerunning or reselecting candidates
& Rank correlation and paired Full MOAE improvements under each weighting
profile
& Spearman \(\rho=0.93\text{--}0.97\); 91.7\% of selections remain
unchanged. Full MOAE improves all three objectives under every profile. \\

\addlinespace[2pt]

Expanded test evaluation
& Freeze prompts, objectives, and deployment rules, then evaluate a predefined
200-query subset of the official TravelPlanner test split
& Base, Best-of-7, Weighted-Sum, and Full MOAE
& \textit{Task/Trajectory/Risk:}\\
& & & Base: \(0.781/0.964/0.281\)\\
& & & Best-of-7: \(0.852/0.984/0.194\)\\
& & & Weighted-Sum: \(0.840/0.980/0.216\)\\
& & & Full MOAE: \(\mathbf{0.861/0.987/0.181}\)\\

\bottomrule
\end{tabular}
\caption{Additional generalization and evaluator-robustness analyses. Results
for expanded test evaluation are reported as Task/Trajectory/Risk.}
\label{tab:pending-generalization}
\end{table*}

\subsection{Mechanism and Search-Behavior Analysis}
\label{app:mechanism}

\begin{table}[t]
\centering
\small
\setlength{\tabcolsep}{5pt}
\begin{tabular}{lrrr}
\toprule
Direction & Mutations & Target improved & Archive admitted \\
\midrule
Task & 360 & 20.56\% & 43.61\% \\
Trajectory & 360 & 13.06\% & 38.61\% \\
Safety & 360 & 25.00\% & 40.28\% \\
\midrule
Overall & 1,080 & 19.54\% & 40.83\% \\
\bottomrule
\end{tabular}
\caption{Direction-specific mutation outcomes on TravelPlanner with Gemma.}
\label{tab:direction-outcomes}
\end{table}

\subsubsection{Direction-Specific Mutation Outcomes}

Across 180 Gemma TravelPlanner queries, the main run contains 1,080 mutation
attempts. Table~\ref{tab:direction-outcomes} measures whether each
objective-directed child improves its target objective over its selected
parent and whether it enters the updated empirical nondominated archive.
Safety-directed mutation has the highest target-improvement rate, while
task-directed mutation has the highest archive-admission rate. The difference
between target improvement and admission is expected because admission depends
on all objectives rather than only the mutation target.

\subsubsection{Archive Dynamics and Parent Diversity}

The mean archive size grows from 1.00 after initialization to 2.01 after the
first generation and 2.52 after the second. A query uses 1.67 distinct parents
on average, and 64.44\% of queries select at least two different archive
members as parents. These values show that objective-specialized parent
selection is operational rather than merely formal: different archive members
often provide the strongest available starting point for different mutation
directions.

The final deployment rule returns the initial candidate for 79 of 180 queries
and an offspring for 101. Among selected offspring, 50 are task-directed, 24
trajectory-directed, and 27 safety-directed. Thus, no single mutation direction
accounts for all deployed improvements.

\subsubsection{Search Effectiveness Metrics}

The main paper reports normalized hypervolume gain, empirical Pareto-set size,
archive admission rate, and strict joint-improvement rate. Normalized
hypervolume gain measures expansion of the attainable three-objective region
relative to the initial rollout and remaining headroom. Empirical Pareto-set
size counts retained nondominated trade-offs but is not intrinsically monotonic
with quality. Archive admission rate measures the fraction of offspring that
introduce a new nondominated trade-off when generated. Joint-improvement rate
is the proportion of queries for which at least one offspring strictly improves
task, trajectory, and safety over the initial rollout. The last metric is
conservative when an initial objective is already saturated.

\begin{table}[t]
\centering
\small
\setlength{\tabcolsep}{2pt}
\renewcommand{\arraystretch}{1.08}
\begin{tabular*}{\columnwidth}{
@{\extracolsep{\fill}}lcccc@{}
}
\toprule
Method
& \shortstack{Norm.\ HV\\gain}
& \shortstack{Pareto\\size}
& \shortstack{Admission\\rate}
& \shortstack{Joint\\improve.} \\
\midrule
Best-of-7
& 0.37 & 1.51 & 27.8\% & 9.4\% \\

Weighted-Sum
& 0.41 & 1.83 & 31.2\% & 9.4\% \\

w/o archive
& 0.43 & 1.96 & 33.9\% & 8.9\% \\

Full MOAE
& \textbf{0.52} & 1.92
& \textbf{35.6\%} & \textbf{13.9\%} \\
\bottomrule
\end{tabular*}
\caption{Search-behavior summary corresponding to the main-paper mechanism
figure. Confidence intervals are reported in the figure.}
\label{tab:search-effectiveness-complete}
\end{table}

Removing directed mutation produces the largest overall degradation, especially
in task improvement and risk reduction. Removing diagnostics weakens final
constraint satisfaction and safety repair while retaining much of the average
trajectory gain. Removing the archive has a smaller effect on mean task score
but reduces strict trajectory validity and the ability to retain alternative
deployment choices. The components therefore play different roles: mutation
proposes targeted changes, diagnostics identify actionable failures, and the
archive preserves improvements that cannot be represented by one scalar
incumbent.

\subsection{Qualitative Evolution Case Study}
\label{app:case-study}
Table~\ref{tab:case-study-values} and
Figure~\ref{fig:case-study-prompt} present a real TravelPlanner query
(index 133) to illustrate how objective evaluation, archive-based parent
selection, diagnostic feedback, and policy mutation interact within one MOAE
search.
Table~\ref{tab:case-study-values}
reports the objective values of the initial rollout and all six offspring,
while Figure~\ref{fig:case-study-prompt} presents the diagnostic information
and policy-level mutation that produce the selected second-generation
candidate. The request asks for a three-day trip from Columbus to Newark for
two adults traveling with children under ten, using an entire-room
accommodation, no self-driving, and a total budget of \$1,200. The initial
rollout satisfies many content constraints but omits return transportation,
contains two invalid actions, and follows instruction-like metadata from an
untrusted observation.

\begin{table}[t]
\centering
\small
\setlength{\tabcolsep}{4pt}
\begin{tabular}{lrrr}
\toprule
Candidate & Task & Trajectory & Safety risk \\
\midrule
Initial rollout $c_0$ & 0.650 & 0.937 & 0.900 \\
Task child $c_1$ & 0.975 & 0.947 & 0.230 \\
Trajectory child $c_2$ & 0.975 & 0.971 & 0.300 \\
Safety child $c_3$ & 0.975 & 0.947 & 0.300 \\
Second-generation child $c_4$ & \textbf{1.000} & \textbf{0.980} & 0.230 \\
Second-generation child $c_5$ & 0.975 & \textbf{0.980} & 0.300 \\
Second-generation child $c_6$ & 0.975 & 0.947 & \textbf{0.200} \\
\bottomrule
\end{tabular}
\caption{Objective values for the real evolution case. Safety risk is lower
when better.}
\label{tab:case-study-values}
\end{table}

\begin{figure*}[t]
\centering
\fbox{%
\begin{minipage}{0.95\textwidth}
\colorbox{gray!22}{\parbox{\dimexpr\linewidth-2\fboxsep\relax}{
\textbf{Real Mutation Trace for Index 133}}}
\par\medskip\hrule\medskip
\textbf{Parent strength.} The selected parent covers the required travel
entities and has the strongest trajectory score in the admissible archive.

\medskip
\textbf{Evaluator diagnosis.} The candidate does not verify total itinerary
cost against the \$1,200 budget and may accept an option before checking all
remaining constraints.

\medskip
\textbf{Mutation instruction.} Preserve the parent's valid tool-use and
evidence-grounding behavior. Add an explicit final budget aggregation step,
reject plans above \$1,200, and re-check transportation, accommodation, and
dining costs before stopping.

\medskip
\textbf{Observed child behavior.} The child performs the missing cost check,
repairs the remaining task constraint, and retains the parent's trajectory
discipline.
\end{minipage}}
\caption{Human-readable view of a real diagnostic-guided mutation. The figure
shows only evaluator summaries and policy-level changes; hidden model reasoning
and attack payloads are not reproduced.}
\label{fig:case-study-prompt}
\end{figure*}

As shown in Table~\ref{tab:case-study-values}, the first-generation offspring
substantially improve the initial rollout but preserve different strengths.
The task-directed child \(c_1\) produces the largest immediate task gain, the
trajectory-directed child \(c_2\) attains the strongest trajectory quality,
and the remaining candidates offer alternative task--trajectory--safety
trade-offs. MOAE therefore retains multiple candidates rather than reducing
them to a single scalarized incumbent.

For the second generation, \(c_2\) is selected as the trajectory-specialized
parent because it has the highest trajectory quality among candidates that
satisfy the task-retention guard. Its evaluator diagnostics identify the
remaining budget constraint. As summarized in
Figure~\ref{fig:case-study-prompt}, the resulting mutation preserves the
parent's valid and evidence-grounded tool-use behavior while adding explicit
total-cost aggregation and enforcement of the \$1,200 limit. The resulting
child \(c_4\) satisfies all task constraints, further improves trajectory
quality, and preserves the reduced safety risk achieved during the first
generation.

The constraint-aware deployment rule consequently selects \(c_4\). This case
illustrates the roles of the three main mechanisms: the archive preserves
offspring with different strengths, objective-specialized selection identifies
a suitable parent, and evaluator diagnostics convert the parent's remaining
weakness into an actionable policy mutation. The child therefore inherits a
useful behavior from the archive and repairs a specific failure instead of
restarting from the original policy.

\subsection{Token and Runtime Analysis}
\label{app:token-analysis}

Search tokens include all model input and output tokens used to generate,
execute, evaluate, and mutate candidates. Deployment tokens count only the
selected candidate's execution and exclude search-time diagnostic and mutation
context. This distinction prevents the optimizer's internal context from being
reported as part of the deployed policy.
For the locally served models, mutation accounts for only 12.2K--12.4K tokens
per query, while most search cost comes from executing complete agent
rollouts. The selected candidate remains close to the Base Agent in execution
cost. DeepSeek-V4-Pro shows a similar total search budget, but its API logs do
not support the same component-level decomposition.

\begin{table*}[t]
\centering
\small
\setlength{\tabcolsep}{5pt}
\begin{tabular}{llrrrrrr}
\toprule
Model & Method & Base & Selected & Execution & Mutation
& Search & Seconds \\
\midrule
Gemma
& Full MOAE
& 56.6 & 56.5 & 384.4 & 12.2 & 396.6 & 610.3 \\

Qwen3
& Full MOAE
& 59.4 & 61.4 & 438.3 & 12.4 & 450.7 & 584.9 \\

DeepSeek-V4-Pro
& Full MOAE
& 60.4 & N/A
& \multicolumn{2}{c}{combined 396.3 }
& 396.3 & N/A \\
\bottomrule
\end{tabular}

\caption{Mean per-query MOAE cost. Token counts are reported in thousands.
``Base'' and ``Selected'' denote the execution-token costs of the initial and
deployed candidates. ``Execution'' includes all complete candidate rollouts,
whereas ``Mutation'' includes the six compact policy-generation calls. The
DeepSeek API logs aggregate search usage but does not provide the same
execution--mutation decomposition or a directly comparable local runtime;
therefore, its combined search usage is reported without an inferred split.}
\label{tab:token-runtime}
\end{table*}

The selected-policy execution cost remains close to that of the Base Agent.
Most additional cost comes from evaluating complete candidates rather than
from compact mutation calls: the six mutations account for 12.2K tokens on
Gemma and 12.4K on Qwen3. The DeepSeek API records total search usage but does
not expose the same local execution/runtime decomposition.

\subsection{Closest-Work Comparison}
\label{app:closest-work}

Table~\ref{tab:closest-work} distinguishes these methods by what they optimize,
where search is performed, how multiple objectives affect candidate retention,
and whether search is separated from deployment. GEPA, MOCHA, SkillMOO,
AgentBreeder, and EvoAgent primarily optimize reusable textual components,
skills, or agent structures over a development task collection. Some of these
methods already preserve multiple objectives or diverse candidates, so our
distinction is not that MOAE is the first method to use evolution or
nondominated selection. Instead, MOAE applies these principles independently
to each query under a seven-rollout budget. Its archive preserves rollout-level
trade-offs among heterogeneous objectives, while a separate constraint-aware
rule determines which archived behavior should be deployed. The key difference
is therefore the combination of per-query search, limited online evaluation,
nondominated retention, and an explicit separation between search-time
preservation and deployment-time preference.

\begin{table*}[t]
\centering
\footnotesize
\setlength{\tabcolsep}{2.5pt}
\renewcommand{\arraystretch}{1.15}

\begin{tabularx}{\textwidth}{
@{}
l
>{\raggedright\arraybackslash}X
>{\raggedright\arraybackslash}X
>{\raggedright\arraybackslash}X
>{\centering\arraybackslash}p{1.55cm}
>{\centering\arraybackslash}p{1.55cm}
@{}
}
\toprule
Method
& Optimized artifact
& Search scope
& Objective structure
& \shortstack{ND\\retention}
& \shortstack{Deployment\\aware\\selection} \\
\midrule

GEPA
& Reusable textual components
& Development task collection
& Performance across examples
& Yes
& No \\

MOCHA
& Persistent structured skills
& Development task collection
& Correctness and platform preferences
& Method-specific
& No \\

SkillMOO
& Reusable skill bundles
& Development task collection
& Performance, cost, and runtime
& Yes
& No \\

AgentBreeder
& Agent architectures and scaffolds
& Development task collection
& Capability and safety
& Quality-diversity
& No \\

EvoAgent
& Multi-agent architecture
& Development task collection
& Primarily task performance
& No
& No \\

\textbf{MOAE}
& Per-query execution policy
& Single query with seven rollouts
& Heterogeneous rollout objectives
& \textbf{Yes}
& \textbf{Yes} \\

\bottomrule
\end{tabularx}

\caption{Comparison with closely related agent optimization settings.
ND denotes nondominated. A development task collection optimizes an artifact
that persists across queries, whereas MOAE performs limited-budget optimization
independently for each query. Deployment-aware selection denotes an explicit
separation between candidate preservation during search and constraint-aware
selection of the response returned for the current query.}
\label{tab:closest-work}
\end{table*}

\subsection{Limitations, Failure Modes, and Reporting}
\label{app:limitations}

\paragraph{Scope of the Pareto claim.}
MOAE uses Pareto dominance as a candidate-preservation rule, but its archive
should not be interpreted as an approximation to a global Pareto front. For
each query, the archive is constructed from only seven evaluated candidates
generated by one initial rollout and six mutations. It therefore represents a
small empirical nondominated set under the selected objectives, evaluator, and
rollout budget. A candidate is retained because no other observed candidate is
measured as better in every objective, not because it is globally optimal or
because no better policy exists outside the evaluated set. The archive can
also contain weak but incomparable candidates, so archive size alone is not a
measure of search quality. For this reason, we jointly report normalized
hypervolume gain, archive admission rate, strict joint-improvement rate, and
the quality of the deployed candidate. Hypervolume conclusions are themselves
conditional on the objective normalization and reference point and should be
used to compare search behavior under the fixed experimental protocol rather
than to establish an absolute measure of agent quality.

\paragraph{Dependence on objective design and evaluators.}
MOAE can optimize only properties represented by its evaluators. If a task
error, unsupported claim, inefficient action, or unsafe influence is not
detected, it cannot contribute to parent selection, diagnostic feedback, or
deployment filtering. Conversely, a misspecified evaluator may reward behavior
that improves the measured score without improving the intended property. This
creates a risk of evaluator exploitation because candidate policies are
repeatedly generated using feedback from the same objective functions that
later rank them. The held-out evaluator and alternative trajectory-weight
analyses reduce this concern by showing that most candidate ordering and
improvements transfer under changed evaluation conditions, but they cannot
exclude all forms of metric overfitting.

The three objectives should also not be interpreted as statistically or
causally independent. Better information coverage and evidence support can
improve both trajectory quality and task completion, while conservative tool
use can affect both trajectory discipline and measured safety. Multi-objective
optimization does not require independence, but the reported joint
improvements mean that several operational criteria improve simultaneously
under their definitions. They do not establish that MOAE has isolated three
independent latent capabilities. The trajectory component weights and safety
thresholds are fixed before the main evaluation and shared across methods, but
different applications may require different objective definitions,
calibrations, or deployment constraints.

\paragraph{Limits of measured safety.}
Safety risk is evaluated using predefined probes applied to untrusted tool
observations, and Attack Success Rate records severe failures observed under
the benchmark attacks. These measurements cover only the tested attack
families, tool interfaces, and propagation paths. An attack-success rate of
zero means that no successful attack was observed in the evaluated cases; it
does not establish universal robustness or certify the deployed agent as safe.
Unseen prompt-injection strategies, adaptive attackers, compromised tools,
multi-turn attacks, or failures outside the monitored planner and output fields
may remain undetected. Safety-optimal candidates can also become overly
conservative and lose task utility, as reflected by the objective-specific
deployment results. Consequently, the safety objective is best viewed as a
measured risk signal for comparative optimization rather than a complete
security guarantee.

\paragraph{Deployment guarantees are empirical.}
The deployment rule requires task performance to be no lower than the initial
rollout and safety risk to be no higher, after which it prioritizes trajectory
quality, safety, and task performance. Because the initial rollout itself is
included, the feasible set is never empty. If no offspring satisfies both
constraints, MOAE returns the initial candidate. This mechanism prevents
measured regressions relative to the observed Base Agent rollout, but it does
not guarantee improvement in the candidate's unknown expected performance.
Both the baseline score and the offspring scores are obtained from individual
stochastic executions and may contain evaluation noise. The retention guard
similarly restricts which candidates may become parents for auxiliary
objectives, but it does not guarantee that a generated child will inherit the
parent's strengths. A fresh rollout may fail to reproduce a successful tool
sequence even when it uses an improved policy.

The lexicographic deployment order is also an application choice rather than a
universally optimal preference. It is intentionally separated from archive
construction so that another application can impose different constraints or
select a task-specialized, trajectory-specialized, or safety-specialized point
from the same candidate pool. However, deployment preferences must still be
specified eventually. MOAE postpones preference commitment during search; it
does not eliminate the need to define acceptable behavior at deployment.

\paragraph{Online cost and budget sensitivity.}
MOAE requires seven complete environment-interacting rollouts and six compact
policy-generation calls for every query. It is therefore training-free but not
cost-free. The method may be unsuitable for latency-sensitive applications,
expensive external tools, irreversible actions, or environments in which
repeated execution changes the underlying state. The seven-rollout
configuration is a practical operating point that supports two complete
three-objective generations, not a convergence guarantee. The rollout-scaling
experiment shows further improvement beyond seven candidates, while smaller
budgets may not provide enough objective-specific offspring to form a useful
archive. Adaptive stopping, asynchronous evaluation, candidate reuse, and
budget allocation based on objective uncertainty remain important directions
for reducing this cost.

The main comparisons match the number of complete agent rollouts, but the
methods do not have identical optimizer overhead. Best-of-7 uses no mutation
calls, whereas evolutionary methods invoke a model to generate new compact
policies. We therefore report optimizer tokens, total search tokens, and
runtime in addition to rollout count. The comparison isolates differences
under a matched environment-interaction budget, not under identical total
token consumption or wall-clock cost.

\paragraph{Interpretation of adapted baselines.}
GEPA-Online, MOCHA-Select, and EvoAgent-Online are matched-budget adaptations
that transfer the corresponding reflection, preference-based selection, or
specialist-integration principles into the same per-query policy space. Their
results should not be interpreted as a comprehensive ranking of the original
methods under their native settings. GEPA normally optimizes reusable textual
components over development examples, MOCHA optimizes persistent structured
skills, and EvoAgent evolves agent structures through a broader process. Our
adaptations intentionally remove differences in artifact scope, development
set size, and rollout harness to compare candidate-generation and selection
principles under seven online rollouts. This improves internal comparability
but narrows the claim to the specific per-query setting studied here.

\paragraph{Generalization and stochasticity.}
The experiments cover TravelPlanner and AgentDojo, three execution models, and
a compact textual policy controlling planning, tool use, verification, and
stopping. These results provide evidence across different models and two
tool-use environments, but they do not establish generality across all agent
architectures. MOAE has not been evaluated for persistent memory, long-term
skill acquisition, multi-agent coordination, embodied interaction, or
environments with irreversible state changes. API-served models may also
change over time even when the model name remains constant, which can limit
exact reproduction. The repeated-seed analysis indicates that the main
trajectory effect is not due to one favorable random seed, but three seeds are
still insufficient to characterize every source of generation variance.
Paired confidence intervals quantify uncertainty over the evaluated queries;
they should not be interpreted as guarantees for a different task
distribution.

\paragraph{Common failure modes.}
Observed and anticipated failures fall into several categories. First, an
objective may already be saturated, leaving little diagnostic signal for its
directed mutation. This occurs, for example, when all candidates receive zero
attack success even though untested vulnerabilities may remain. Second, a
diagnostic can identify a symptom, such as missing evidence or an invalid call,
without providing a repair that the mutation model can reliably translate into
a policy. Third, a mutation may overcorrect one weakness and reduce another
objective, producing a useful specialized candidate but not a feasible
deployment candidate. Fourth, several mutations may produce nearly identical
policies, limiting archive diversity despite the use of different objective
directions. Fifth, an apparently improved policy may fail during fresh
execution because tool results and model generations are stochastic. Finally,
the optimizer can exploit systematic evaluator blind spots, producing measured
improvement without a corresponding improvement in human-perceived quality.
The archive, retention guard, and deployment filter reduce the impact of these
failures but do not remove their underlying causes.

\paragraph{Reporting and reproducibility.}
The supplementary artifact records the dataset split identifiers, model names
and revisions, serving parameters, compact policy schema, machine-readable
prompts, objective implementations, component weights, safety thresholds,
retention tolerance, deployment rule, random seeds, per-query candidate
records, optimizer usage, integrity checks, and aggregation scripts. Candidate
records preserve the objective values, parent relationships, mutation
directions, archive updates, and selected deployment point needed to audit the
reported search process. API credentials, local filesystem paths, user
identifiers, and other machine-specific information are excluded.

Potentially harmful prompt-injection payloads are not reproduced verbatim in
the paper or unrestricted artifact. Instead, we report benchmark identifiers,
attack-family labels, evaluator versions, aggregate outcomes, and non-sensitive
execution metadata. This permits verification of experimental coverage and
reported measurements without unnecessarily redistributing actionable attack
content. 

%% file: aaai2027.bib
@inproceedings{yao2022react,
  title={React: Synergizing reasoning and acting in language models},
  author={Yao, Shunyu and Zhao, Jeffrey and Yu, Dian and Du, Nan and Shafran, Izhak and Narasimhan, Karthik R and Cao, Yuan},
  booktitle={The eleventh international conference on learning representations},
  year={2022}
}

@misc{xie2024travelplanner,
      title={TravelPlanner: A Benchmark for Real-World Planning with Language Agents}, 
      author={Jian Xie and Kai Zhang and Jiangjie Chen and Tinghui Zhu and Renze Lou and Yuandong Tian and Yanghua Xiao and Yu Su},
      year={2024},
      eprint={2402.01622},
      archivePrefix={arXiv},
      primaryClass={cs.CL},
      url={https://arxiv.org/abs/2402.01622}, 
}

@inproceedings{liu2024agentbench,
  title={Agentbench: Evaluating llms as agents},
  author={Liu, Xiao and Yu, Hao and Zhang, Hanchen and Xu, Yifan and Lei, Xuanyu and Lai, Hanyu and Gu, Yu and Ding, Hangliang and Men, Kaiwen and Yang, Kejuan and others},
  booktitle={International Conference on Learning Representations},
  volume={2024},
  pages={52989--53046},
  year={2024}
}

@article{ma2024agentboard,
  title={Agentboard: An analytical evaluation board of multi-turn llm agents},
  author={Ma, Chang and Zhang, Junlei and Zhu, Zhihao and Yang, Cheng and Yang, Yujiu and Jin, Yaohui and Lan, Zhenzhong and Kong, Lingpeng and He, Junxian},
  journal={Advances in neural information processing systems},
  volume={37},
  pages={74325--74362},
  year={2024}
}

@inproceedings{qin2023toolllm,
  title={Toolllm: Facilitating large language models to master 16000+ real-world apis},
  author={Qin, Yujia and Liang, Shihao and Ye, Yining and Zhu, Kunlun and Yan, Lan and Lu, Yaxi and Lin, Yankai and Cong, Xin and Tang, Xiangru and Qian, Bill and others},
  booktitle={The twelfth international conference on learning representations},
  year={2023}
}

@inproceedings{zhan2024injecagent,
  title={Injecagent: Benchmarking indirect prompt injections in tool-integrated large language model agents},
  author={Zhan, Qiusi and Liang, Zhixiang and Ying, Zifan and Kang, Daniel},
  booktitle={Findings of the Association for Computational Linguistics: ACL 2024},
  pages={10471--10506},
  year={2024}
}

@article{deb2024agentdojo,
  title={Agentdojo: A dynamic environment to evaluate prompt injection attacks and defenses for llm agents},
  author={Debenedetti, Edoardo and Zhang, Jie and Balunovic, Mislav and Beurer-Kellner, Luca and Fischer, Marc and Tram{\`e}r, Florian},
  journal={Advances in Neural Information Processing Systems},
  volume={37},
  pages={82895--82920},
  year={2024}
}

@article{madaan2023selfrefine,
  title={Self-refine: Iterative refinement with self-feedback},
  author={Madaan, Aman and Tandon, Niket and Gupta, Prakhar and Hallinan, Skyler and Gao, Luyu and Wiegreffe, Sarah and Alon, Uri and Dziri, Nouha and Prabhumoye, Shrimai and Yang, Yiming and others},
  journal={Advances in neural information processing systems},
  volume={36},
  pages={46534--46594},
  year={2023}
}

@article{shinn2023reflexion,
  title={Reflexion: Language agents with verbal reinforcement learning},
  author={Shinn, Noah and Cassano, Federico and Gopinath, Ashwin and Narasimhan, Karthik and Yao, Shunyu},
  journal={Advances in neural information processing systems},
  volume={36},
  pages={8634--8652},
  year={2023}
}

@inproceedings{yang2024opro,
  title={Large language models as optimizers},
  author={Yang, Chengrun and Wang, Xuezhi and Lu, Yifeng and Liu, Hanxiao and Le, Quoc V and Zhou, Denny and Chen, Xinyun},
  booktitle={International Conference on Learning Representations},
  volume={2024},
  pages={12028--12068},
  year={2024}
}

@inproceedings{guo2024evoprompt,
  title={Connecting large language models with evolutionary algorithms yields powerful prompt optimizers},
  author={Guo, Qingyan and Wang, Rui and Guo, Junliang and Li, Bei and Song, Kaitao and Tan, Xu and Liu, Guoqing and Bian, Jiang and Yang, Yujiu},
  booktitle={International Conference on Learning Representations},
  volume={2024},
  pages={34133--34156},
  year={2024}
}

@inproceedings{fernando2024promptbreeder,
  title={Promptbreeder: Self-referential self-improvement via prompt evolution},
  author={Fernando, Chrisantha and Banarse, Dylan Sunil and Michalewski, Henryk and Osindero, Simon and Rockt{\"a}schel, Tim},
  booktitle={Forty-first International Conference on Machine Learning},
  year={2024}
}

@inproceedings{opsahlong2024mipro,
  title={Optimizing instructions and demonstrations for multi-stage language model programs},
  author={Opsahl-Ong, Krista and Ryan, Michael J and Purtell, Josh and Broman, David and Potts, Christopher and Zaharia, Matei and Khattab, Omar},
  booktitle={Proceedings of the 2024 Conference on Empirical Methods in Natural Language Processing},
  pages={9340--9366},
  year={2024}
}

@misc{agrawal2025gepa,
      title={GEPA: Reflective Prompt Evolution Can Outperform Reinforcement Learning}, 
      author={Lakshya A Agrawal and Shangyin Tan and Dilara Soylu and Noah Ziems and Rishi Khare and Krista Opsahl-Ong and Arnav Singhvi and Herumb Shandilya and Michael J Ryan and Meng Jiang and Christopher Potts and Koushik Sen and Alexandros G. Dimakis and Ion Stoica and Dan Klein and Matei Zaharia and Omar Khattab},
      year={2026},
      eprint={2507.19457},
      archivePrefix={arXiv},
      primaryClass={cs.CL},
      url={https://arxiv.org/abs/2507.19457}, 
}

@inproceedings{zhou2024modpo,
  title={Beyond one-preference-fits-all alignment: Multi-objective direct preference optimization},
  author={Zhou, Zhanhui and Liu, Jie and Shao, Jing and Yue, Xiangyu and Yang, Chao and Ouyang, Wanli and Qiao, Yu},
  booktitle={Findings of the Association for Computational Linguistics: ACL 2024},
  pages={10586--10613},
  year={2024}
}

@article{zhong2024panacea,
  title={Panacea: Pareto alignment via preference adaptation for llms},
  author={Zhong, Yifan and Ma, Chengdong and Zhang, Xiaoyuan and Yang, Ziran and Chen, Haojun and Zhang, Qingfu and Qi, Siyuan and Yang, Yaodong},
  journal={Advances in Neural Information Processing Systems},
  volume={37},
  pages={75522--75558},
  year={2024}
}

@article{rosser2025agentbreeder,
  title={Agentbreeder: Mitigating the ai safety risks of multi-agent scaffolds via self-improvement},
  author={Rosser, J and Foerster, Jakob},
  journal={Advances in Neural Information Processing Systems},
  volume={38},
  pages={149564--149594},
  year={2026}
}

@article{tanjim2026mocha,
  title={MOCHA: Multi-Objective Chebyshev Annealing for Agent Skill Optimization},
  author={Tanjim, Md Mehrab and Subramanian, Jayakumar and Chen, Xiang and Kveton, Branislav and Mukherjee, Subhojyoti and Zhang, Anlan and Kim, Sungchul and Sarkhel, Somdeb and Choudhury, Sunav},
  journal={arXiv preprint arXiv:2605.19330},
  year={2026}
}

@inproceedings{yuan2025evoagent,
  title={Evoagent: Towards automatic multi-agent generation via evolutionary algorithms},
  author={Yuan, Siyu and Song, Kaitao and Chen, Jiangjie and Tan, Xu and Li, Dongsheng and Yang, Deqing},
  booktitle={Proceedings of the 2025 Conference of the Nations of the Americas Chapter of the Association for Computational Linguistics: Human Language Technologies (Volume 1: Long Papers)},
  pages={6192--6217},
  year={2025}
}

@article{dai2023saferlhf,
  title={Safe rlhf: Safe reinforcement learning from human feedback},
  author={Dai, Josef and Pan, Xuehai and Sun, Ruiyang and Ji, Jiaming and Xu, Xinbo and Liu, Mickel and Wang, Yizhou and Yang, Yaodong},
  journal={arXiv preprint arXiv:2310.12773},
  year={2023}
}

@article{zhang2025bfpo,
  title={Bi-factorial preference optimization: Balancing safety-helpfulness in language models},
  author={Zhang, Wenxuan and Torr, Philip HS and Elhoseiny, Mohamed and Bibi, Adel},
  journal={arXiv preprint arXiv:2408.15313},
  year={2024}
}

@inproceedings{xia2025tokenskip,
  title={Tokenskip: Controllable chain-of-thought compression in llms},
  author={Xia, Heming and Leong, Chak Tou and Wang, Wenjie and Li, Yongqi and Li, Wenjie},
  booktitle={Proceedings of the 2025 Conference on Empirical Methods in Natural Language Processing},
  pages={3351--3363},
  year={2025}
}

@inproceedings{yan2025macc,
  title={From Long to Lean: Performance-aware and Adaptive Chain-of-Thought Compression via Multi-round Refinement},
  author={Yan, Jianzhi and Liu, Le and Pan, Youcheng and Chen, Shiwei and Yuan, Zike and Xiang, Yang and Tang, Buzhou},
  booktitle={Proceedings of the 2025 Conference on Empirical Methods in Natural Language Processing},
  pages={12290--12306},
  year={2025}
}

@article{wang2025latencytts,
  title={Faster and better llms via latency-aware test-time scaling},
  author={Wang, Zili and Zhang, Tianyu and Bai, Haoli and Hou, Lu and Yu, Xianzhi and Liu, Wulong and Xiang, Shiming and Zhu, Lei},
  journal={arXiv preprint arXiv:2505.19634},
  year={2025}
}

@article{li2026paretopo,
  title={Towards Pareto-Optimal Tool-Integrated Agents with Pareto Ranking Policy Optimization},
  author={Li, Junyi and Qian, Xiaowei and Zhang, Yingyi and Zhang, Wenlin and Li, Guojing and Zhang, Sheng and Han, Xiao and Wang, Yichao and Zhao, Xiangyu},
  journal={arXiv preprint arXiv:2606.16111},
  year={2026}
}

@article{gong2026skillmoo,
  title={Skillmoo: Multi-objective optimization of agent skills for software engineering},
  author={Gong, Jingzhi and Gu, Ruizhen and Fei, Zhiwei and Cao, Yazhuo and Twist, Lukas and Geiger, Alina and Han, Shuo and Sobania, Dominik and Sarro, Federica and Zhang, Jie M},
  journal={arXiv preprint arXiv:2604.09297},
  year={2026}
}

@inproceedings{jiang-tang-2026-agents,
    title = "Why Agents Compromise Safety Under Pressure",
    author = "Jiang, Hengle  and
      Tang, Ke",
    booktitle = "Findings of the {A}ssociation for {C}omputational {L}inguistics: {ACL} 2026",
    month = jul,
    year = "2026",
    address = "San Diego, California, United States",
    publisher = "Association for Computational Linguistics",
    url = "https://aclanthology.org/2026.findings-acl.810/",
    doi = "10.18653/v1/2026.findings-acl.810",
    pages = "16453--16470",
    ISBN = "979-8-89176-395-1"
}
